%% file: jmlr-sample.tex
\documentclass[pmlr,10pt]{jmlr}

\input{preamble/math}

\input{preamble/packages}

\newcommand{\suppfigref}[1]{\hyperref[#1]{Supplementary Figure \ref*{#1}}}
\newcommand{\supptabref}[1]{\hyperref[#1]{Supplementary Table \ref*{#1}}}
\newcommand{\methodsref}[1]{\hyperref[#1]{Methods}}
\newcommand{\supplementref}[1]{\hyperref[#1]{Supplement}}
\usepackage{booktabs}
\usepackage{siunitx}
\usepackage[switch]{lineno}

\newcommand{\equal}[1]{{\hypersetup{linkcolor=black}\thanks{#1}}}

\title[New-Onset Diabetes Assessment Using Artificial Intelligence-Enhanced
Electrocardiography]{New-Onset Diabetes Assessment Using Artificial Intelligence-Enhanced
Electrocardiography}

\author{
\Name{Hao Zhang}\equal{These authors contributed equally} \Email{hz1975@nyu.edu}\\
\Name{Neil Jethani}\footnotemark[1]\equal{Now at Indiana University School of Medicine}
\Email{nj594@nyu.edu}\\
\Name{Aahlad Puli}
\Email{aahlad@nyu.edu}\\
\addr New York University, USA
\AND
\Name{Leonid Garber}
\Email{Leonid.Garber@nyulangone.org}\\
\addr NYU Langone Health, USA
\AND
\Name{Lior Jankelson}
\Email{Lior.Jankelson@nyulangone.org}\\
\addr NYU Langone Health, USA\\
\addr NYU Tandon School of Engineering, USA
\AND
\Name{Yindalon Aphinyanaphongs}
\Email{Yin.A@nyulangone.org} \\
\addr NYU Langone Health, USA
\AND
\Name{Rajesh Ranganath}
\Email{rajeshr@cims.nyu.edu}\\
\addr New York University, USA
}
\begin{document}

\maketitle
\vspace{-0.5in}
\begin{abstract}
Diabetes has a long asymptomatic period which can often remain undiagnosed for multiple years. In this study, we trained a deep learning model to detect new-onset diabetes using 12-lead ECG and readily available demographic information. To do so, we used retrospective data where patients have both a hemoglobin A1c and ECG measured. However, such patients may not be representative of the complete patient population. As part of the study, we proposed a methodology to evaluate our model in the target population by estimating the probability of receiving an A1c test and reweight the retrospective population to represent the general population. We also adapted an efficient algorithm to generate Shapley values for both ECG signals and demographic features at the same time for model interpretation. The model offers an automated, more accurate method for early diabetes detection compared to current screening efforts. Their potential use in wearable devices can facilitate large-scale, community-wide screening, improving healthcare outcomes.
\end{abstract}
\vspace{0.5in}
\paragraph*{Data and Code Availability}
The datasets generated and/or analyzed during the current study are not publicly available due to specific institutional requirements governing privacy protection. The codes can be found in a GitHub repository\footnote{\url{https://github.com/rajesh-lab/a1c_ecg}}.

\paragraph*{Institutional Review Board (IRB)}
This study was approved by the Institutional Review Board (IRB) of NYU Grossman School of Medicine under IRB protocol ID\#i20-00348. 
Informed consent was waived by the Institutional Review Board (IRB) of NYU Grossman School of Medicine for this study. 
All methods employed in the study were conducted in strict accordance with the relevant guidelines and regulations.

\section{Introduction}
\label{sec:intro}

Diabetes mellitus affects 38.4 million Americans, 95\% of whom have type 2 diabetes, and is one of the leading causes of illness in the United States \citep{CDC2020}.
Due to its long asymptomatic period, early diagnosis requires screening, which allows for interventions that reduce diabetic complications \citep{Simmons2017}.
Several tests allow for the discovery of diabetes in asymptomatic people, including glycated hemoglobin (HbA1c), fasting plasma glucose (FPG), and oral glucose tolerance test (OGTT). One advantage that HbA1c offers over other tests is that no fasting is required, better preanalytical stability, and less day-to-day variations \citep{adacare2024}.
Yet, even with current screening efforts, 1 in 5 people with diabetes are not aware of their diabetes status \citep{CDC2020}.
To address this, public health agencies such as the American Diabetes Association (ADA) have promoted the use of risk tests. 
The ADA Risk test uses simple-to-collect patient information, such as family history and body mass index (BMI), to identify high-risk patients \citep{adacare2024}.

Artificial intelligence (AI), however, has created the possibility of improving diabetes assessment by making use of structured data in the electronic records and high-dimensional biosignals like the electrocardiogram (ECG).
ECGs are simple to collect, even in the community setting, because of ECG-enabled mobile fitness trackers.
Further, ECGs are often collected at a visit where labs were not, thus using the ECG alone provides more opportunities to screen for diabetes.
In this study, we trained a deep learning model to classify high HbA1c using a 12-lead/10 second ECG and readily available demographic information
and evaluated its ability to screen for new-onset diabetes in the outpatient population who receives an ECG.
To do so, we used retrospective data where patients have both an HbA1c and ECG measured. 
However, such patients may not be representative of the target population, so we generated a pseudo-population that better represents the target population, as visualized in \figureref{fig:pseudopopulation}.
This procedure is important because the AI-enhanced system must perform well on those eligible for diabetes assessment, such as patients who have an ECG collected, not only those with both contemporaneous ECG and HbA1c measurements.
We have also performed an external validation to demonstrate how the model performs on a population the models have not seen before to evaluate how well the model is able to generalize across populations. 

\input{figures/psuedopopulation}

\section{Related Work}
\paragraph{Diabetes Detection Using AI-enhanced ECG Systems} AI-enhanced ECG systems have performed tasks that are imperceptible by humans, such as identifying age, sex \citep{Attia2019}, and left ventricular systolic dysfunction \citep{Attia2019a}.
Preliminary evidence suggests that a relationship between ECGs and diabetes exists.
A deep learning system using ECG data was able to detect nocturnal hypoglycemic events \citep{Porumb2020}.
Most recently, researchers used an AI-enabled ECG system to estimate HbA1c, where the estimates were associated with many complications of diabetes such as chronic kidney diseases and heart failure \citep{Lin2021}.
A limitation of this previous work is that the population investigated includes many patients with long-standing diabetes, as they are more likely to have their A1c measured.
Patients with long-standing diabetes often develop cardiovascular diseases; therefore, it is less surprising that the ECG can help identify such patients.
Instead, we focus on detecting new-onset diabetes in patients without a history of diabetes.
There have been previous works in machine learning algorithms that predicts new-onset diabetes with structured lab features including triglycerides, cholesterol, and fasting plasma glucose with decent success \citep{mldm_1,mldm_2,mldm_3,mldm_4}.
However, the collection of such lab features required drawing blood and processing by a lab, where an HbA1c can also be computed, might not be accessible to everyone, especially in rural and underdeveloped areas. 

\paragraph{Efficient Shapley Value Estimation with FastSHAP} 
Shapley value is a widely used approach to interpret black-box models, but they are known to be computationally expensive and can be hard to be applied to large, high-dimensional models \citep{vdb2021}.
FastSHAP is a method to estimate Shapley values in a single forward pass using a learned explainer model \citep{jethani2022fastshap}. 
FastSHAP was trained via stochastic gradient descent using a weighted least squares objective function, making the training efficient \citep{jethani2022fastshap}. 
FastSHAP can generate estimations at a fraction of the time compared to other approaches like KernelSHAP, with higher-quality explanations than gradient-based methods \citep{jethani2022fastshap}. 
Previous work has shown that FastSHAP can be adopted to explain a deep learning model that detects right bundle branch block from ECG inputs, but is limited to a single data modality \citep{jethani2023dont}. 
Here, we customized the FastSHAP algorithm to estimate Shapley values for both unstructured (ECGs) and structured (demographics) inputs. 

\section{Methods} \label{sec:methods}
\input{sections/methods}

\section{Results} 
\input{sections/results}

\section{Discussion} \label{sec:discussion}
\input{sections/discussion}

\section*{Acknowledgments}
We wish to thank John Higgins from Massachusetts General Hospital for his feedback. Neil Jethani was partially supported by NIH T32GM007308 and T32GM136573. Yin Aphinyanaphongs was partially supported by NIH 3UL1TR001445-05 and National Science Foundation award \#1928614. This work was partly supported by the NIH/NHLBI Award R01HL148248, NSF Award 1922658 NRT-HDR: FUTURE Foundations, Translation, and Responsibility for Data Science, NSF CAREER Award 2145542, NSF Award 2404476, ONR N00014-23-1-2634, and Optum. We would also like to thank the support by IITP with a grant funded by the MSIT of the Republic of Korea in connection with the Global AI Frontier Lab International Collaborative Research.

\bibliography{sample}

\newpage
\newpage
\appendix
\counterwithin{figure}{section}
\counterwithin{table}{section}
\renewcommand\thefigure{\thesection\arabic{figure}}
\renewcommand\thetable{\thesection\arabic{table}}
\input{sections/supplement} \label{sec:supp}

\end{document}

%% file: preamble/math.tex
\def\eqref#1{equation~\ref{#1}}

\def\1{\bm{1}}

\def\rvm{{\mathbf{m}}}

\def\rvx{{\mathbf{x}}}

\def\rvy{{\mathbf{y}}}
\def\rvz{{\mathbf{z}}}

\DeclareMathAlphabet{\mathsfit}{\encodingdefault}{\sfdefault}{m}{sl}
\SetMathAlphabet{\mathsfit}{bold}{\encodingdefault}{\sfdefault}{bx}{n}

\DeclareMathOperator*{\E}{\mathbb{E}}

%% file: preamble/packages.tex
\usepackage{amssymb, amsmath}
\usepackage{mathtools}
\usepackage{xcolor}

\usepackage{microtype}
\usepackage[english]{babel}
\usepackage[parfill]{parskip}
\usepackage{enumitem}
\usepackage{wrapfig}
\usepackage{multicol}
\usepackage{float}
\usepackage{adjustbox}

\usepackage[T1]{fontenc}
\usepackage{amsfonts}
\usepackage{bm}
\usepackage{bbm}
\usepackage{anyfontsize}

\usepackage{fancyhdr}

\usepackage{graphicx}
\usepackage{wrapfig}
\usepackage{tikz}
\usepackage{forest}
\usetikzlibrary{arrows.meta, arrows, shapes, positioning, decorations.pathreplacing}

\usepackage{booktabs}
\usepackage{array}

\usepackage{listings}
\usepackage{fancyvrb}
\fvset{fontsize=\normalsize}

\usepackage{natbib}

\usepackage[acronym,nowarn]{glossaries}
\glsdisablehyper

\usepackage[all]{hypcap}

\usepackage{nameref}

\usepackage{url}
\usepackage[nameinlink]{cleveref}

\usepackage{multirow}
\usepackage{nicefrac}
\usepackage{todonotes}
\usepackage{placeins}

%% file: figures/psuedopopulation.tex
\begin{figure*}[htbp]
    \floatconts{fig:pseudopopulation}{\caption{\small{\textbf{Diagram of pseudo-population construction.} The electronic health record provides a large amount of data with which to train and evaluate an AI-enhanced ECG to estimate HbA1c. However, for many patients ECGs or HbA1c tests are not performed. In order to understand how well the AI-enhanced ECG will work in practice, one needs to estimate the performance on the complete population. This diagram shows that by modeling the probability of ordering an HbA1c, the observed population can be re-weighted to represent the complete population.}}}
    {\includegraphics[width=0.7\textwidth]{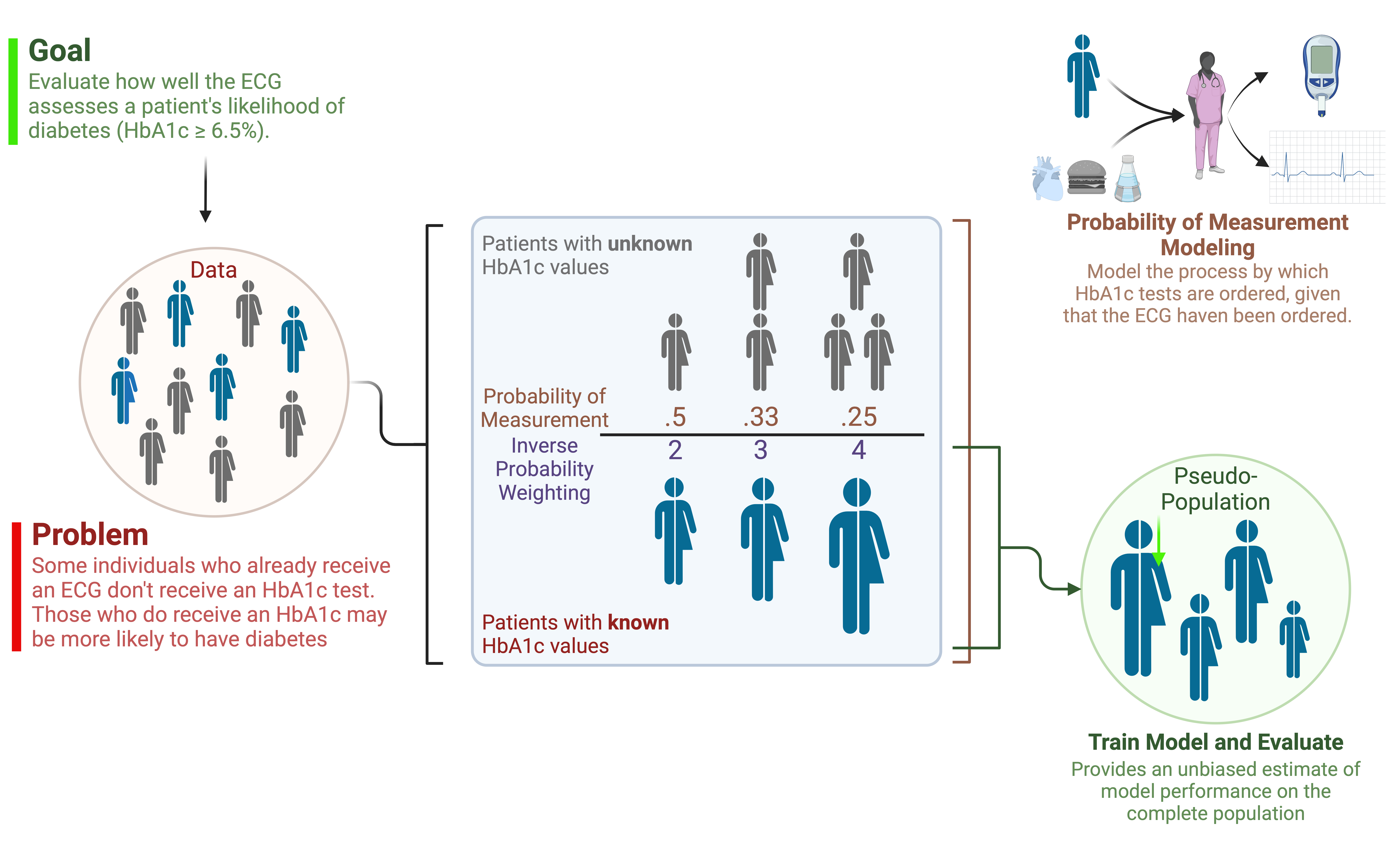}}
\end{figure*}

%% file: sections/methods.tex
\subsection{Study Overview}
Training deep learning models for clinical screening from retrospective data presents a fundamental challenge: our data contains only patients where both the ECG and HbA1c were measured, yet the intended deployment population includes all patients who receive an ECG. 
Physicians selectively order HbA1c tests based on risk factors like age, BMI, and comorbidities, meaning the observed population differs systematically from the target population. Models trained naively on this selected sample may perform poorly when deployed clinically. 
To address this selection bias, we employ inverse probability weighting (IPW) to reweight the observed population to represent the target population (\figureref{fig:pseudopopulation}). Our methodology proceeds in three stages. 
First, we model the probability that a patient receives an HbA1c test given they received an ECG (\Cref{ipa_model}). 
Second, we apply IPW during both model training and evaluation to ensure our model optimizes performance on all ECG recipients, not just those selected for HbA1c testing (\Cref{ipa}). 
Third, we train a deep convolutional neural network that fuses ECG signals with demographic information to detect new-onset diabetes (\Cref{model_arch}). 
While IPW is established in dealing with missing data, its integration into deep learning training and evaluation for clinical AI represents a novel application addressing a critical gap, as most clinical AI studies train on convenience samples without accounting for how data acquisition differs from target deployment. 
To ensure that future evaluations would yield similar results, we examined the sensitivity of our results when incorrect weights were applied under violations of missing-at-random assumption (\Cref{sen}).
Our framework provides a principled solution to this pervasive problem. 
Additionally, we extend the FastSHAP algorithm to generate explanations for multi-modal inputs (ECG signals and tabular features simultaneously), enabling efficient interpretation of our model (\Cref{model_intepretation}).

\subsection{Study Population}
We considered Manhattan, Brooklyn, and Mineola in Long Island outpatient encounters within NYU Langone Health system between January 1, 2013 and September 17, 2021. 
We selected encounters in the study period with HbA1c obtained and a standard 12-lead/10s ECG taken up to 30 days before the encounter.
We then split the cohort into train, validation, and test set with 3:1:1 ratio. 
We further filtered the test set to those without prior history of diabetes, which we defined as the absence of a diagnosis for diabetes (ICD-10 Code E10, E11, or E13) and no HbA1c $\geq$ 6.5\% prior to encounter. 
The test set was used to evaluate the model performance. 
Patient characteristics and p-values were generated using the ``tableone'' Python package \citep{tableone}.

\subsection{Data}  
Encounter data were collected from the electronic health record (EHR) system and ECGs were retrieved from MUSE (GE Healthcare, Chicago, IL).
ECGs were measured at sampling rates of either 250 or 500 Hz.
The signals were filtered using the biosppy Python package with a finite impulse response (FIR) filter and an additional band-pass filter \citep{biosppy}.
For each encounter, we extracted the patient's demographics as well as active problems (from ICD-10 codes), medications, and laboratory results, including their HbA1c values, prior to the encounters (\supptabref{tab:variables}). 
Diagnoses covered risk factors for diabetes/cardiovascular disease, disease complications, or medications for the treatment of diabetes/cardiovascular diseases. 
Variables with missing values were imputed by multivariate imputation with chained equations using the scikit-learn package \citep{scikit-learn}.

\subsection{Probability of Measurement Model} \label{ipa_model}
To model the mechanism of data acquisition, we sampled one outpatient encounter per patient in the study period. 
We trained an XGBoost model to classify whether or not an HbA1c were acquired given the patient had an ECG measured. 
A single encounter was randomly selected for each patient in the study cohort to train the IPW model, yielding 483,289 encounters.
The patient characteristics of the data set can be found in \supptabref{tab:outpatient_table1}.
The full set of variables summarized in \supptabref{tab:variables} that would inform a physician's decision to acquire an HbA1c was considered and labeled the data with whether an HbA1c were obtained. 
Any variables with missing values were imputed to mimic a clinician's ability to make decisions with the available information about a patient.  
We tuned the XGBoost algorithm's hyperparameters and selected the best performing model (the one with the highest area under the precision-recall curve (AUPRC)) using 5-fold cross-validation. 

\subsection{Inverse Probability Adjustment}  \label{ipa}
The probability of acquiring an HbA1c was obtained by running each sample through the probability of measurement model.
To reduce the variability in downstream inverse probability weighted estimates, the most extreme probabilities were truncated to the [0.02-0.98] range.
Expanding this range to [0.005-.995], where less than 10\% of the probabilities are truncated, did not affect the estimated model performances.
Inverse probability weighting (IPW) re-weighs each encounter by the inverse probability of acquiring an HbA1c.
To ensure equal emphasis across all patients for which predictions would be made, patients that, for example, were less likely to have the HbA1c measured were up-weighted \citep{Horvitz01121952}.
The inverse probability weighted estimator is described in detail in \nameref{prob_meas}.
Further, for training, each encounter was down-weighted for patients with multiple encounters in the dataset by dividing by the number of encounters to ensure that each patient had equal representation in the dataset.

\subsection{Model Architecture} \label{model_arch}
We implemented a convolutional neural network (CNN) to learn a concise 1-dimensional representation of the ECG time series.
This representation was fused with the tabular data then fed through a fully-connected neural network with a softmax output layer to generate the probability of each class. 
We selected tabular features that are routinely collected in clinical settings without additional testing and that are established diabetes risk factors.
Such tabular features include: age, sex, race, ethnicity, and BMI.
The CNN architecture was based on the current state of the art for arrhythmia detection and is a 34-layer ResNet CNN consisting of 16 residual connections as depicted in \suppfigref{fig:model_architecture} \citep{Hannun2019}.
The input to the network was an 8 x 2,500 matrix, representing the 8 measured leads (lead III and the augmented leads are arithmetically computed) by 10-second duration sampled at 250 Hz (ECGs sampled at 500Hz were down-sampled to 250 Hz).
Model training details can be found in \Cref{supp_training}.

\subsection{Model Evaluation} \label{model_evaluation}
The evaluation task is new-onset diabetes detection. 
The test set consists of 28,907 encounters involving 25,951 patients without prior history of diabetes, of which 4.9\% have an HbA1c $\geq$6.5\%. 
\autoref{tab:data_split_table1} describes the patient characteristics.
We labeled each encounter by binarizing whether the HbA1c measured was $\geq 6.5\%$, indicating new-onset diabetes.
Accordingly, for each encounter, we obtained the scores outputted by the model, and summed the scores under bins with HbA1c $\geq$ 6.5\%.
We then re-weighted each encounter by their inverse probability of ECG/HbA1c measurement. 
We constructed the receiver-operator (ROC) and precision-recall (PRC) curves and calculated the area under the ROC (AUC) and PRC (AURPC). 
We set up two baselines for diabetes screening: 1) the ADA Risk test, 2) QDiabetes-2018 \citep{qdb}.
We did not include additional features added to 2018 version of the QDiabetes risk score because prior validations showed similar performance to prior version of the risk score \citep{qdb}.
QDiabetes-2018 scores were generated using the ``QDiabetes'' R package.

To evaluate each model against the ADA Risk Test, we binarized the ADA Risk Test score at each level from 1 to 7 and computed the true positive rate (TPR).
We then calculated model thresholds that correspond to each ADA Risk Test score TPR, which is used to binarized the model output. 
We finally assessed the positive predictive value (PPV) at each TPR for all models. This evaluation quantifies the difference in false alarms.
As the ADA recommends that people with ADA risk score $\geq 5$ should follow up with their doctor, we paid particular attention to the performance across models at this level.
For each model, the threshold that matched the TPR of the ADA risk score when the ADA risk score is $\geq 5$ was defined as the high-risk threshold for that model. 

To assess its impact on current clinical practice, we looked at those patients that have no history of diabetes 
. 
We first classified these patients for new-onset diabetes using the model predictions at the high-risk threshold, and then used the ECG model's PPV at this threshold to estimate how many additional patients have diabetes but were missed during screening. 
The idea is that if testing rates are low in patients with a high estimated likelihood of diabetes, where the likelihood is indicated by the PPV, then there is an opportunity to improve the quality of care. 

For all evaluation metrics in this section, we adopted the bootstrap method to calculate confidence intervals as well as p-values when comparing models. 
We bootstrapped for 1000 rounds for each metric, thus p-values have a resolution of \(\frac{1}{1001}\).

\subsection{Sensitivity Analysis} \label{sen}
The inferences in patients with ECG measured rely on assumptions about the mechanism controlling the acquisition/missingness of HbA1c data -- that they are missing at random (MAR) given the set of observed risk factors (\supptabref{tab:variables}).
Violations of this assumption may arise due to recording errors in the electronic health record or other bits of patient information that goes unrecorded.
They can be reflected in the degree to which the probability for inverse weighting differs using the recorded information from the true probability that would render the data MAR.

Therefore, we examined how the results change when significant violations to the MAR assumptions were made. 
We considered the following two violations: 1) removing age from the probability of measurement model, and 2) ignore the weight completely (unweighted). 
The first violation simulated a scenario where the clinicians order HbA1cs without considering the patient's age. 
The second violation indicated that the HbA1cs were ordered at random. 
We thresholded each model using the high-risk threshold, and compared the AUC and PPV across models for each reweighting scheme. 

\subsection{Prospective Analysis}
We hypothesized that the false positives identified by the model may still present a higher risk of developing diabetes in the future. 
To evaluate the model on a longer time horizon, we followed up patients within the test set who did not develop diabetes at the time, and collected any HbA1c measurements or diabetes diagnoses within one year since the original measurements. 
To test the hypothesis, patients were categorized into high-risk and low-risk groups according to the ECG model's and ADA Risk Test's respective thresholds. 
We then generated Kaplan-Meier curves and compared the cumulative incidence of future diabetes in one year between high-risk and low-risk groups based on only the model thresholds.
For the analysis, we treated the comparisons as two separate analyses, one for each model. 
The curves were generated using the lifelines Python package. Confidence intervals were calculated using Greenwood's exponential formula and log-rank tests were used to compute p-values. 

\subsection{Model Interpretations} \label{model_intepretation}
We calculated Shapley values to provide explanations for how each of tabular feature and the ECG signal influence predictions. 
To efficiently calculate Shapley values, we adopted FastSHAP algorithm and customized the code to generate explanations for tabular features and ECG signal at the same time. 
The algorithm is a two-step process \citep{jethani2022fastshap}.
We first trained a surrogate model that learns to generate predictions with masked input, then an explainer model which generates Shapley values using the surrogate model. 
The superpixel size is 1 for tabular features, and 25 for ECG signals, which corresponds to 0.1 seconds of ECG signal. 
All eight leads were considered at once when calculating the Shapley values for the ECG signal. 
Details on FastSHAP can be found in \Cref{supp_fastshap}.

We generated a violin plot for the tabular features to plot Shapley values for each data point and colored by the feature value, where red means high feature value and blue means low feature value. 
For ECG signal, we selected three samples from the most positively-scored by the our ECG model, and overlaid top 20\% absolute Shapley values for each sample, colored with the same palette as the tabular features. 

%% file: sections/results.tex
\subsection{Model Performance}
The ECG model, which takes in age, sex, race, ethnicity, BMI, and ECG, achieved the best performance (\tableref{tab:model_metrics},  AUC, 0.80 [95\% CI, 0.79-0.80]), outperforming scores based on using the ADA Risk test (AUC, 0.68 [95\% CI, 0.67-0.69]), and QDiabetes-2018 AUC, 0.70 [95\% CI, 0.69-0.70]). 
Comparing positive predictive value (PPV) across models, matching the TPR of ADA $\geq 5$ (TPR 0.72), the ECG model showed significantly superior precision (PPV, 0.11 [95\% CI, 0.11-0.12]) than ADA Risk Test (PPV, 0.08 [95\% CI, 0.07-0.08]) and QDiabetes-2018 (PPV, 0.08 [95\% CI, 0.07-0.08]). The PPV for the ADA Risk Test is consistent with prior assessment of the ADA Risk Test \citep{Bang2009}. The p-values < 0.01 for all comparisons. 
This indicates that at the same detection rate as ADA risk test, ECG model generated significantly less false alarms than the baselines. 
The AUROC curve, PPV at all ADA Risk Test scores, and additional results can be found in \Cref{supp:add_results}.

\begin{table*}[!ht]
    \centering
    \begin{tabular}{lccc}
    \hline
        Metric & ECG Model & ADA Risk Test & QDiabetes-2018 \\
    \hline
        Recall (TPR) & 0.72 & 0.72 & 0.72 \\
        Precision (PPV) & \textbf{0.11 [0.11--0.12]} & 0.08 [0.07--0.08] & 0.08 [0.07--0.08] \\
        F1-Score & \textbf{0.19 [0.18--0.20]} & 0.14 [0.13--0.15] & 0.14 [0.13--0.15] \\
        AUC & \textbf{0.80 [0.79--0.80]} & 0.68 [0.67--0.69] & 0.70 [0.69--0.70] \\
    \hline
    \end{tabular}
    \caption{\textbf{Model performance on test set} Matching the TPR of ADA $\geq 5$ (TPR 0.72), the ECG model outperformed ADA Risk Test and QDiabetes-2018 across metrics including PPV, F1-score, and AUC. }
    \label{tab:model_metrics}
\end{table*}

Using the high-risk threshold, we evaluated how well the ECG model can automatically screen patients that receive an ECG.
In the clinical utility set, 112,403 encounters (103,865 patients) where an ECG was acquired for patients without a prior history of diabetes.
13.6\% of the set had both an ECG and an HbA1c measured, leaving with 87,183 patients were not assessed with an HbA1c test. 
The ECG model identified 36,922 of these patients as highly likely to have diabetes.
We found that the ECG model outperformed (PPV, 0.14 [95\% CI, 0.13-0.14]) the ADA risk test (PPV, 0.08 [95\% CI, 0.07-0.08]) and QDiabetes (PPV, 0.09 [95\% CI, 0.08-0.10]). 
The estimated PPV implies that 14\% of this cohort or 4,501 additional patients would have been newly diagnosed with diabetes had the ECG model been applied clinically to this cohort.

\subsection{Prospective Analysis}
Kaplan-Meier curves on the subset of the test set who did not develop diabetes by the time of the encounter were illustrated in  \figureref{fig:prospective}. 
Both high-risk groups identified by the ADA risk test and the ECG model presented significantly higher risk than the low risk groups (p-value <0.001). 
However, the high-risk group by the ECG model presented a higher, 4.5-fold increase in cumulative incidence rate over the corresponding low-risk group, compared to the ADA risk test, which had only a 2.7-fold increase (p-value < 0.001). 
That is, the ECG model is better at separating those who have high future high-risk from low-risk than ADA risk test. 
\input{figures/prospective}
\subsection{Sensitivity Analysis} 
\figureref{fig:sensitivity_auc} and \figureref{fig:sensitivity_ppv} compares the AUC and PPV of the ECG model to the baselines when significant violations were made to the MAR assumptions. 
The ECG model significantly (p-value $<0.01$) outperformed the baselines both when the age was taken out of consideration or the model completely unweighted. 
The results showed that even under impossible situations like ordering HbA1c without considering age or ordering randomly, the ECG model is still superior over the baselines. We have also conducted additional sensitivity analysis that can be found in \Cref{add_sen_analysis}. 

\subsection{Model Interpretation}
Shapley values were generated both for the tabular inputs were shown in \figureref{fig:shap_tab}.
The tabular features were ranked by their average absolute Shapley values.
Samples of the most positively-scored ECGs and their corresponding Shapley values can be seen in \figureref{fig:shap_beat} (ECGs are periodic and contain redundant information across beats).
We have also performed additional analysis on the explanations generated by the Shapley value algorithm (see \Cref{supp_fastshap}), where we have shown that by only using a few ECG features identified as important by the Shapley values, we were able to recover 50\% of the performance supplied by the ECG waveform. 

\input{figures/screening_plots}

\input{figures/fastshap}

%% file: figures/prospective.tex
\begin{figure}[htbp]
    \floatconts{fig:prospective}{\caption{\small{\textbf{Kaplan-Meier analysis.}} High-risk group identified by the ECG Model showed significant higher cumulative incidence compared to the high-risk group identified by the ADA Risk Test.}}
{\includegraphics[width=.425\textwidth]{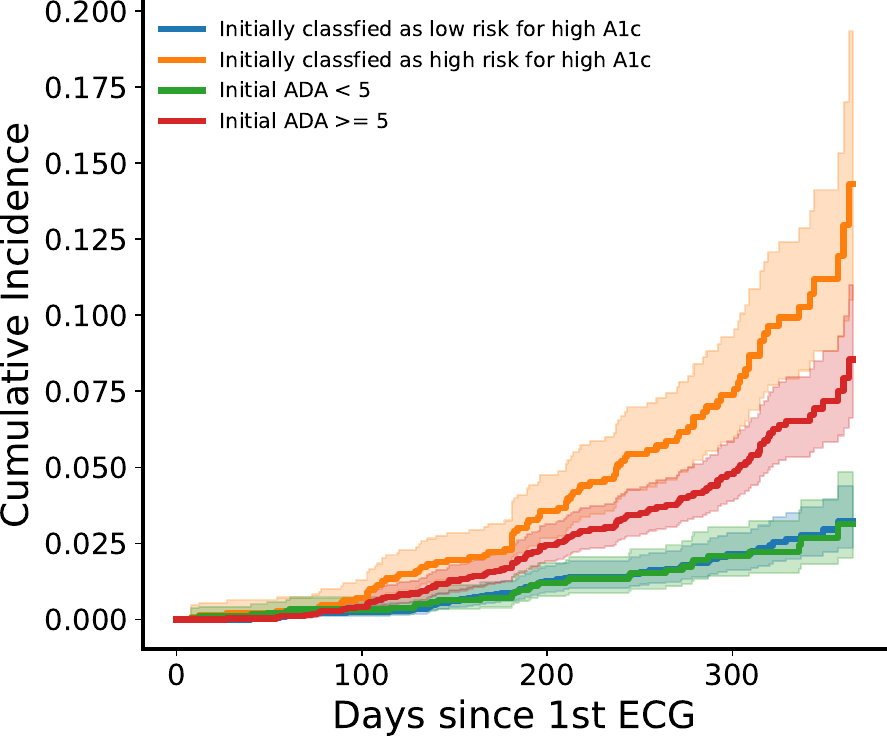}}
\end{figure}

%% file: figures/screening_plots.tex
\begin{figure}[htb]  %
  \floatconts{fig:screening}{\caption{\small{\textbf{Sensitivity analysis.} Even with significant violations made to MAR, the ECG model still outperformed both baselines. }}}{
    \subfigure[AUC with different weighting schemes]{\label{fig:sensitivity_auc}\includegraphics[width=.35\textwidth]{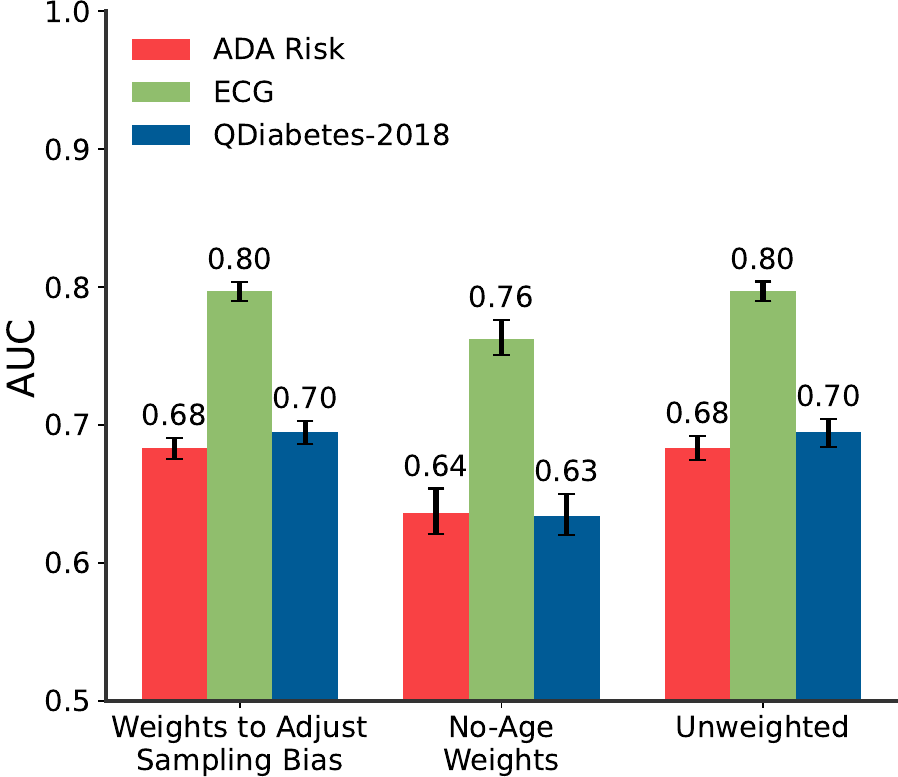}}
   \subfigure[PPVs with different weighting schemes]{\label{fig:sensitivity_ppv}\includegraphics[width=.35\textwidth]{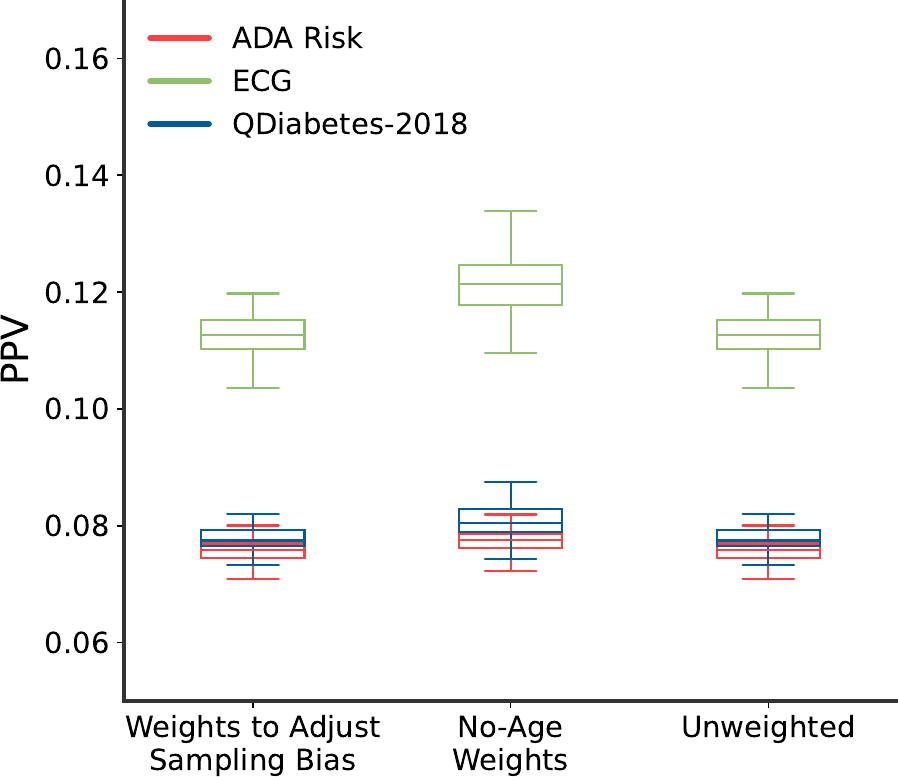}} 
        }
\end{figure}

%% file: figures/fastshap.tex
\begin{figure*}[htbp]
    \floatconts{fig:fastshap}{\caption{\small{\textbf{Shapley values quantifying feature contributions to model predictions.}} \textbf{(a)} Distribution of Shapley values for demographic features across test set, ranked by mean absolute value. Each point is one patient; color indicates feature value (red=high, blue=low). Positive values increase predicted diabetes probability. \textbf{(b)} Shapley values for three highest-scoring ECG samples. Top 20\% absolute values (0.1s superpixels, all leads) overlaid on Lead I. Red segments increase predicted probability; blue decrease it.}}
{    \subfigure[Tabular features and their Shapley Values\label{fig:shap_tab}]{\includegraphics[width=0.4\textwidth]{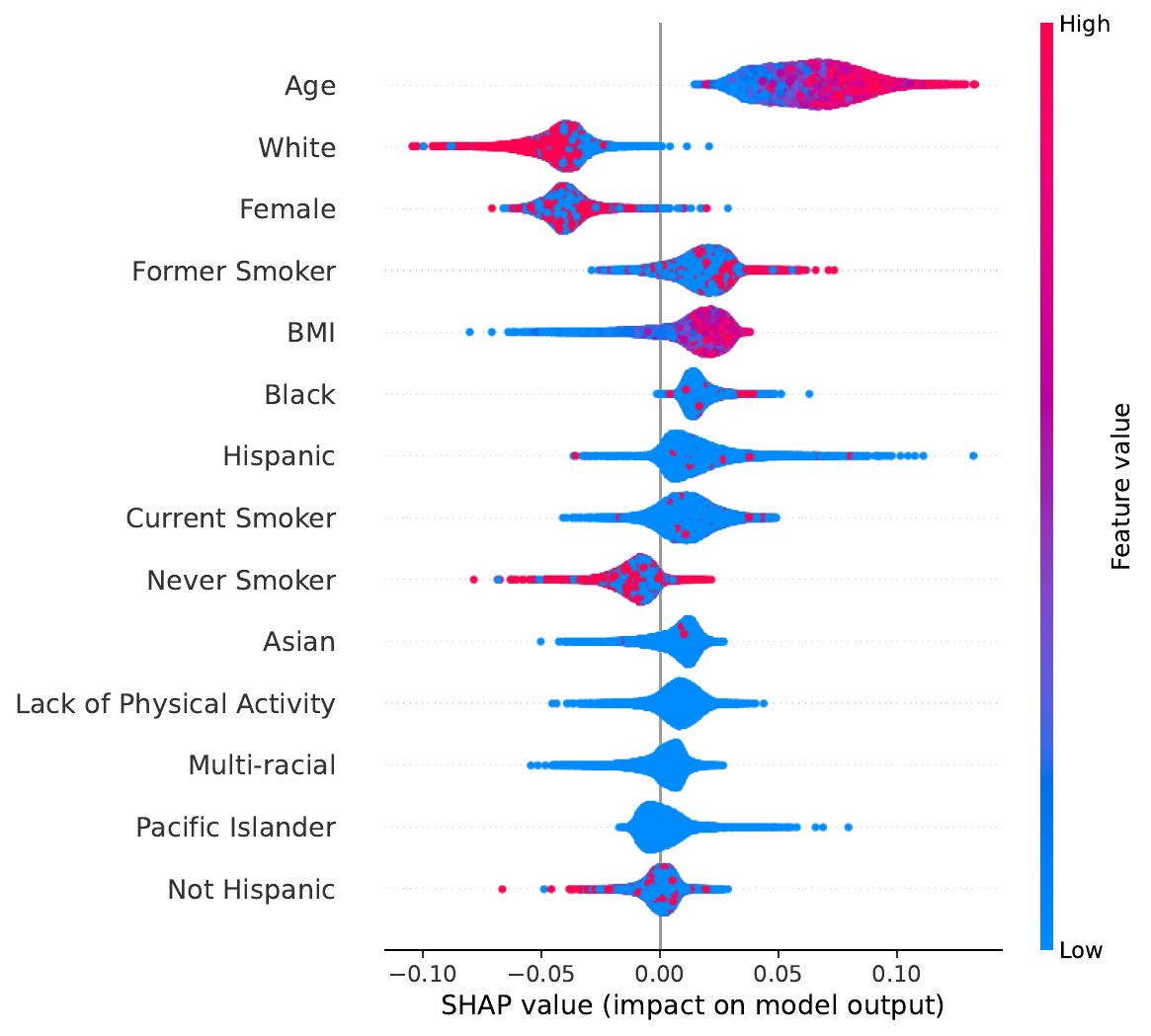}}
    \subfigure[Samples ECG Shapley values\label{fig:shap_beat}]{\includegraphics[width=0.4\textwidth]{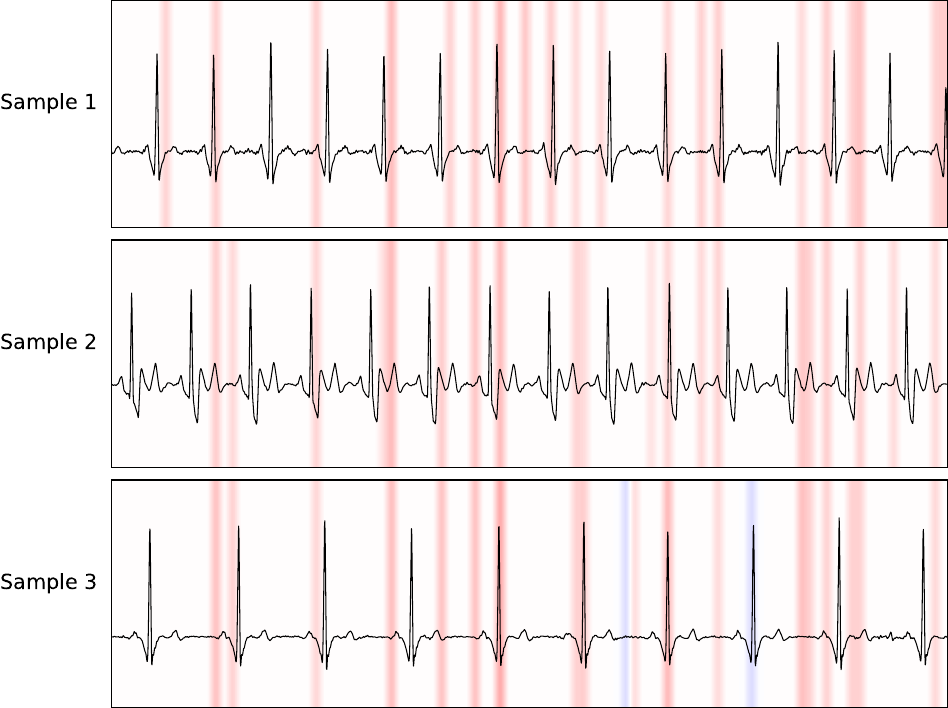}} 
     }
\end{figure*}

%% file: sections/discussion.tex
Undiagnosed diabetes leads to increased morbidity and a higher overall burden on the healthcare system \citep{harris2000early, dall2019economic}.
The capacity to screen for diabetes is primarily limited by both the frequency of visits and the time available at those visits. 
Improving this capacity requires better use of existing patient interactions in clinical settings and an improved capacity to screen for diabetes in the community.
Such improvements will be crucial for identifying the 8.5 million US adults with undiagnosed type 2 diabetes \citep{CDC2020}.

This work demonstrates how diabetes assessment benefits from the incorporation of ECGs. 
We have shown that ECG-based assessment outperforms existing assessment tools, achieving a higher PPV at the desired TPR (0.72).
The higher PPV implies that using the ECG reduces false alerts relative to the ADA risk score.
At the same PPV level (0.08), the ECG had far fewer missed cases than the ADA risk score.
In the outpatient settings, the AI-enhanced model can be run in the background and identify patients with diabetes who had their ECGs taken. If a patient has an ECG taken, the model scores can be then incorporated into routine practice. As 87\% of high-likelihood patients did not have their HbA1c assessed, the ECG model would improve the quality of clinical care and screening for diabetes by detecting cases that would otherwise be missed.

Community-wide AI-enhanced ECG diabetes assessment can be carried out via mobile devices that collect single-lead (Lead-I) ECGs.
We found that an AI-enabled single-lead ECG (Lead I), while it is inferior to the 12-lead ECG model, retains much of the discriminative performance (AUC, 0.78 [95\% CI, 0.77-0.79] vs. 0.80 [95\% CI, 0.79-0.80]) (\suppfigref{fig:single_lead_auroc}). 
These findings suggest that an AI-enabled ECG system could be utilized both at the community level and within outpatient clinics to automate the identification of patients who are likely to have diabetes.
With the improved capacity for diabetes screening enabled by the ECG-based assessment, more diabetes would be captured and the rate of undiagnosed diabetes would go down. 

Based on our prospective validation, even if the patient does not have diabetes at the time of inference, those with scores that were above the high-risk threshold showed a greater chance of developing diabetes in one year, and is superior relative to ADA risk test. 
This suggests that the model can act not only as a screening tool, but a prediction tool as well.
In practice, the model can aide clinicians to decide whether to order more frequent monitoring and/or interventions for these high-risk patients. 
We have also shown that the model is able to maintain performance on a different population the model has not seen before, demonstrating the generalizability of the model. 

We have considered incorporating FGP and OGTT in addition to HbA1c. However, because the volume of FGP/OGTT being ordered in ambulatory setting is much smaller than that of HbA1c, and FGP/OGTT subjects to more day-to-day variations compared to HbA1c, we eventually opted for using HbA1c only. We have also considered the task as a regression model by estimating the HbA1c directly, then classify the patient based on the estimated value. The performance is statistically worse than classification (AUC, 0.78 [95\% CI, 0.77-0.79], p-value $< 0.01$). 
When training the model, we have also found that having four bins also allowed for easier training compared to binarizing HbA1c at the diabetic threshold. 

A strength of our study lies in our methodology for training a model to screen for diabetes in yet unseen populations, recently described as integrative modeling \citep{Hofman2021}.
Practitioners require a large amount of data to train deep learning models.
The ideal training and validation is a complete set of data, however collecting 
a complete set of data requires a new study, and these studies demand prior evidence.
Therefore, studies rely on retrospective study designs.
Researchers look for patient encounters in a retrospective dataset where the input variable (i.e., ECGs) can be paired with an outcome variable (i.e., HbA1c) \citep{Attia2019, Lin2021}.
Yet, it is likely that patient risk factors influence the measurement of these variables. 
This influence means evaluations under the retrospective population, where both the input and output are known, do not match evaluations under the complete population of interest. 

In our work, we looked at the test set representative of the population that would receive an HbA1c measurement using IPW.
This procedure ensured that we optimized over the intended population during training and evaluated our models’ ability to screen the general outpatient population for diabetes. 
Next, we evaluated the sensitivity of our results when incorrect weights were applied under violations to MAR assumptions. 
These steps help ensure that further prospective evaluation on a complete population would yield similar results. 
The framework we lay out for handling missing data using IPW during model training and evaluation will be of use in other domains that make use of retrospective data to build AI models \citep{mitani2020detection, hughes2021deep}.

Finally, using FastSHAP, we were able to draw insights from the ECG model. 
Previous work has shown that FastSHAP can be adopted to explain a deep learning model that detects right bundle branch block from ECG inputs \citep{jethani2023dont}.
This is the first application of the explanation algorithm on a multi-modal model that takes in both structured and unstructured data. 

\subsection{Limitations}
Despite inverse probability weighting to address selection bias, unmeasured confounding and residual bias may remain. While there is no patient crossover between test set and external set, the retrospective, single-institution design may limit generalizability, as data completeness, coding accuracy, and patient characteristics can differ across settings. Prospective, multi-institutional studies are needed to validate the model’s real-world performance and clinical impact. 

\subsection{Conclusion}
Our proposed model significantly outperforms existing screening tools for detecting new-onset diabetes. Applied clinically, our model could identify an estimated 4,501 additional patients with undiagnosed diabetes who were not assessed with HbA1c testing. A key methodological contribution is our inverse probability weighting framework that addresses selection bias in retrospective EHR data, ensuring models generalize to target deployment populations. This approach provides a template for clinical AI research. As ECG acquisition becomes increasingly accessible through wearable devices, our model offers scalable, automated diabetes screening beyond traditional clinical settings, improving early detection and enabling timely intervention.

%% file: sections/supplement.tex
\section{Selected Model Architecture}
\input{figures/model_architecture}
\newpage
\section{List of Covariates}
\input{tables/covariates}
\newpage
\section{Probability of Measurement Modeling} \label{prob_meas}
\subsection{Inverse Probability Weighted Estimator}
Let $\rvy \in \{0,1\}$ be a categorical variable indicating that a patient's HbA1c is $\leq$ 6.4\%, $\geq$ 6.5\%.
Let $\rvx$ represent a patient's ECG, $\rvz$ represent a patient's clinical factors driving the acquisition of an HbA1c given that an ECG was ordered, and $\rvm \in \{0,1\}$ indicate whether or not an HbA1c were acquired. 
HbA1c is observed only when acquired by a physician, indicated that access to samples $\rvy,\rvx,\rvz \sim p(\rvx,\rvy,\rvz \mid \rvm = 1)$ are available.
However, we want to estimate the model performance (the mean of some loss function $f$) on the complete distribution $p(\rvx,\rvy,\rvz)$; $\E_{\rvx,\rvy,\rvz\sim p(\rvx,\rvy,\rvz)}\left[ f(\rvx,\rvy) \right]$.
To achieve this using samples from only from $p(\rvx,\rvy,\rvz \mid m = 1)$, we express the expectation on the complete distribution as a weighted expectation on the observed distribution.

First, as $f(\rvx, \rvy)$ does not depend on $\rvm$,
$\E_{\rvx,\rvy,\rvz\sim p(\rvx,\rvy,\rvz)}\left[ f(\rvx,\rvy) \right] = \E_{\rvx,\rvy,\rvz,\rvm \sim p(\rvx,\rvy,\rvz,\rvm)}\left[ f(\rvx,\rvy) \right]$.
The latter expectation can be expanded using the law of total expectation when conditioning on the missingness indicator $\rvm$ as follows:
\begin{align*}
    \begin{split}
    \E_{\rvx,\rvy,\rvz,\rvm \sim p(\rvx,\rvy,\rvz,\rvm)}\left[ f(\rvx,\rvy) \right] &= \E_{\rvx,\rvy,\rvz \sim p(\rvx,\rvy,\rvz \mid m=1)}\left[ f(\rvx,\rvy) \mid m = 1 \right]p(m = 1) +\\
    &\quad\ 
    \E_{\rvx,\rvy,\rvz \sim p(\rvx,\rvy,\rvz \mid m=0)}\left[ f(\rvx,\rvy) \mid m = 0 \right]p(m = 0)
    \end{split}
\end{align*}
The first expectation on the right hand side is with respect to the observed data distribution and therefore can be directly estimated using the observed data.
Focusing instead on the second expectation on the left hand side, we make the MAR assumption that $\rvy, \rvx \perp \rvm \mid \rvz $ and expand it as follows:
{\small
\begin{align*}
    &\E_{\rvx,\rvy,\rvz \sim p(\rvx,\rvy,\rvz \mid m=0)}\left[ f(x,y) \mid m = 0 \right]p(m = 0) \\
    &= \E_{\rvz\sim p(\rvz \mid m = 0)}\left[\E_{\rvx,\rvy \sim p(\rvx,\rvy \mid m=0, \rvz)}\left[ f(\rvx,\rvy) \mid m = 0, \rvz \right]p(m = 0)\right]\\
    &= \E_{\rvz\sim p(\rvz \mid m = 0)}\left[\E_{\rvx,\rvy \sim p(\rvx,\rvy \mid \rvz)}\left[ f(\rvx,\rvy) \mid \rvz \right]p(m = 0)\right]\\
    &= \E_{\rvz\sim p(\rvz \mid m = 0)}\left[\E_{\rvx,\rvy \sim p(\rvx,\rvy \mid \rvz)}\left[ \frac{p(m=1 | z)p(\rvz)}{p(m=1 | z)p(\rvz)} f(\rvx,\rvy) \mid z \right]p(m = 0)\right]\\
    &= \int \E_{\rvx,\rvy \sim p(\rvx,\rvy \mid \rvz)}\left[ \frac{p(m=1 \mid \rvz)p(\rvz)}{p(m=1 \mid \rvz)p(\rvz)} f(\rvx,\rvy) \mid z \right]p(m = 0)p(\rvz \mid m = 0) d\rvz \\
   &= \int \E_{\rvx,\rvy \sim p(\rvx,\rvy \mid \rvz)}\left[\frac{p(m = 0)p(\rvz \mid m = 0)}{p(m=1 \mid \rvz)p(\rvz)} f(\rvx,\rvy) \mid \rvz \right]p(m=1 \mid \rvz)p(\rvz) d\rvz \\
   &= \E_{\rvz\sim p(\rvz \mid m = 1)}\left[\E_{\rvx,\rvy \sim p(\rvx,\rvy \mid \rvz)}\left[ \frac{p(m = 0 \mid \rvz)}{p(m=1 \mid \rvz)} f(\rvx,\rvy) \mid \rvz \right]p(m = 1)\right]\\
   &= \E_{\rvx,\rvy,\rvz \sim p(\rvx,\rvy,\rvz \mid m=1)}\left[ \frac{p(m = 0 \mid \rvz)}{p(m=1 \mid \rvz)} f(\rvx,\rvy) \right]p(m = 1)\\
\end{align*}
}

Combining the two terms results in the following:
{\small
\begin{align*}
    &\E_{\rvx,\rvy,\rvz,\rvm \sim p(\rvx,\rvy,\rvz,\rvm)}\left[ f(x,y) \right] \\
    =& \E_{\rvx,\rvy,\rvz \sim p(\rvx,\rvy,\rvz \mid m=1)}\left[ f(x,y) \mid m = 1 \right]p(m= 1) \\
    &+ \E_{\rvx,\rvy,\rvz \sim p(\rvx,\rvy,\rvz \mid m=1)}\left[ \frac{p(m = 0 \mid \rvz)}{p(m=1 \mid \rvz)} f(\rvx,\rvy) \right]p(m = 1)\\
    =& \E_{\rvx,\rvy,\rvz \sim p(\rvx,\rvy,\rvz \mid m=1)}\left[ \left(1 + \frac{p(m = 0 \mid \rvz)}{p(m=1 \mid \rvz)}\right) f(\rvx,\rvy) \right]p(m = 1)\\
    =& \E_{\rvx,\rvy,\rvz \sim p(\rvx,\rvy,\rvz \mid m=1)}\left[ \left(\frac{p(m = 1 \mid \rvz)}{p(m=1 \mid \rvz)} + \frac{p(m = 0 \mid \rvz)}{p(m=1 \mid \rvz)}\right) f(\rvx,\rvy) \right]p(m = 1)\\
    =& \E_{\rvx,\rvy,\rvz \sim p(\rvx,\rvy,\rvz \mid m=1)}\left[ \frac{1}{p(m=1 \mid \rvz)} f(\rvx,\rvy) \right]p(m = 1)\\
\end{align*}
The expectation can also be computed with respect to the whole population by constructing a random variable that is zero when there is missing data ($\rvm=0$). With $\mathbbm{1}$ as the indicator function:
\begin{align*}
    \E&_{\rvx,\rvy,\rvz \sim p(\rvx,\rvy,\rvz \mid m=1)}\left[ \frac{1}{p(m=1 \mid \rvz)} f(\rvx,\rvy) \right]p(m = 1) 
    \\
    =& \E_{\rvx,\rvy,\rvz \sim p(\rvx,\rvy,\rvz \mid m=1)}\left[ \frac{1}{p(m=1 \mid \rvz)} f(\rvx,\rvy) \right] (p(m = 1) \mathbbm{1}(1 = 1) +  p(\rvm = 0) \mathbbm{1}(0 = 1))\\    
    =& \E_{\rvx,\rvy,\rvz \sim p(\rvx,\rvy,\rvz \mid m=1)}\left[ \frac{1}{p(m=1 \mid \rvz)} f(\rvx,\rvy) \right] (p(m = 1) \mathbbm{1}(1 = 1)) 
    \\
    &+ \E_{\rvx,\rvy,\rvz \sim p(\rvx,\rvy,\rvz \mid m=1)}\left[ \frac{1}{p(m=1 \mid \rvz)} f(\rvx,\rvy) \right] p(m = 0) \mathbbm{1}(0=1)
    \\
    =& \E_{\rvx,\rvy,\rvz \sim p(\rvx,\rvy,\rvz \mid m=1)}\left[ \frac{1}{p(m=1 \mid \rvz)} f(\rvx,\rvy) \right] (p(m = 1) \mathbbm{1}(1 = 1)) 
    \\
    &+ \E_{\rvx,\rvy\rvz \sim p(\rvx,\rvy,\rvz \mid m=0)}\left[ \frac{1}{p(m=1 \mid \rvz)} f(\rvx,\rvy) \right] p(m = 0) \mathbbm{1}(0 = 1)
    \\
    =& \E_{\rvx,\rvy,\rvz \sim p(\rvx,\rvy,\rvz,\rvm)}\left[ \frac{ \mathbbm{1}(\rvm = 1)}{p(m=1 \mid \rvz)} f(\rvx,\rvy) \right]
\end{align*}
}
The estimator above is known as the inverse probability weighted (IPW) estimator. 
We parametrically model $p(\rvm = 1 \mid z)$ and use the IPW estimator in order to estimate the expected performance in the complete population using only the data observed.

\input{figures/calibration_curve}

\subsection{Additional Sensitivity Analysis} \label{add_sen_analysis}
In addition to the comparison to the two violations, we defined a range over which the probabilities are subject to change and examined how the results change in the worst-case scenario, using an approach inspired by \cite{Zhao}.
To compare the models, we thresholded each model using the TPR corresponding to the ADA Risk Test score of $\geq$ 5.
We then minimized the difference in F1-score, the harmonic mean of the PPV and TPR, between the ECG model and the baseline models. 
To minimize this difference, all the encounters where the ECG model produced an incorrect classification were up-weighted, while the correctly classified encounters were down-weighted.
We defined the quantity that minimized the different in F1-score as the adversarial weight. 

We then calculated the ratio of false positives (FP) to true positives (TP), weighted by the average normalized weight for each classification. We then compare the following weighting schemes: original weights to adjust sampling bias, no-age weighting, adversarial weighting, and unweighted. 
For a list of weights $\textbf{w}$, lists of weights of the true positives and false positives $\textbf{w}_{TP}, \textbf{w}_{FP}$, the average weight $\overline{\textbf{w}}$, and the number of weights in $\textbf{w}$ $|\textbf{w}|$. The ratio of false positives to true positives $TP/FP$ is:
$$
FP/TP = \frac{\overline{\textbf{w}_{FP}}\times|\textbf{w}_{FP}|}{\overline{\textbf{w}_{TP}}\times|\textbf{w}_{TP}|}
$$
The idea is that as these performance metrics like AUC and PPV depend on the true positives and false positives, the higher the ratio, the more influence false positives have on these metrics. 
By comparing these metrics, we can then put the adversarial weight in context.

\input{figures/additional_sensitivity}

\figureref{fig:screening_sensitivity} depicted that in the worst-case scenario, where the F1-score of the ECG model is selectively minimized, the important place on each incorrectly classified encounter would have to be increase by a factor of 1.35 for each incorrectly identified encounter and decrease by a factor of 1.35 for each correctly identified encounter to invalidate the results. 
To put this in context, \tableref{tab:additional_sensitivity_table} details the TP/FP ratio for each weighting scheme. 
A ratio > 1 means the PPV were lowered.
Adversarial weights had a ratio that is higher than both no-age weights and unweighted, indicating that such reweighting schemes are implausible.

\begin{table*}
\floatconts{tab:additional_sensitivity_table}{\caption{\textbf{FP/TP for each weighting scheme}. Adversarial weights had a ratio that is higher than both no-age weights and unweighted, indicating that such reweighting schemes are implausible.}}{
\begin{tabular}{@{}lcccc@{}}
\toprule
 &
  \textbf{\begin{tabular}[c]{@{}c@{}}Weights to Adjust\\ Sampling Bias\end{tabular}} &
  \textbf{\begin{tabular}[c]{@{}c@{}}No-Age\\ Weights\end{tabular}} &
  \textbf{Unweighted} &
  \textbf{\begin{tabular}[c]{@{}c@{}}Adversarial\\ Weights\end{tabular}} \\ \midrule
\multicolumn{1}{r}{\begin{tabular}[c]{@{}r@{}}FP/TP\end{tabular}} &
  8.15 &
  9.21 &
  8.15 &
  14.33
\end{tabular}
}
\end{table*}

\section{FastSHAP Algorithm}\label{supp_fastshap}
As discussed in \nameref{sec:methods}, we opted for FastSHAP algorithm to explain model predictions. 
To accommodate both tabular inputs and ECG inputs, modifications were made based on the Tensorflow version of the FastSHAP code \citep{jethani2022fastshap}.
Here we lay out the changes made for this task. 
\subsection{Surrogate Model}
We took the already-trained ECG model as our basis for the surrogate model used it to score subsets of inputs for each instance \citep{jethani2022fastshap,jethani2021have}.

As the tabular inputs and ECG inputs have different dimensions, we first tiled the tabular inputs to match shape of the ECG inputs, and concatenated the two into a single tensor.
The single tensor was then connected to a custom layer to generate a single mask that can randomly mask portions of both inputs.
Based on the mask, tabular features that were masked had their values replaced with $-1$, and $0$ for the ECG features. 
The masked tensor was then split up back into tabular and ECG inputs, and were fed into their respective portion of the model illustrated in \figureref{fig:model_architecture}.
We used the same training and validation to train this model as the ECG model. 
The model was trained using an Adam optimizer up to 100 epochs to minimize the categorical cross-entropy loss. 
A learning rate scheduler was put in place that multiplies the learning rate by 0.8 after two epochs of no validation loss improvement, and early stopping was trigger after the validation loss did not improve for five epochs. 
The initial learning rate was $1e^{-3}$ and batch size was 128.
The model was trained for a total of 42 epochs before early stopping. 

\subsection{Explainer Model}
Once we have the surrogate model trained, we then train the explainer model, another neural network, that generates Shapley values. This network is $\phi_{\text{fast}}$.
We define a value function:
$$v_{\rvx}(s) = \log(\rvy = \text{pos} | p_{\text{surr}}(m(\rvx,s)),$$
which is the natural logarithm of surrogate model score of the positive class given a masked input $m(\rvx,s)$, where $\rvx$ is the input and $s$ is the mask. 

The inputs of the explainer model were masked in similar fashion as the surrogate model, and the rest of the model architecture follows the original FastSHAP explainer implementation. 
We used additive efficiency normalization to adjust Shapley value predictions.
The model was trained using an Adam optimizer up to 200 epochs to minimize the mean squared error FastSHAP loss. 
The same learning rate scheduler mentioned earlier was put in place for training the explainer model. 
The initial learning rate was $1e^{-5}$ and batch size was 32.
The model was trained for a total of 13 epochs before early stopping. 
Finally, the Shapley values $\phi_{\text{fast}}^{\text{eff}}$ equals to \citep{jethani2022fastshap}:
$$
\phi_{\text{fast}}^{\text{eff}}(\rvx;\theta)= \phi_{\text{fast}}(\rvx;\theta) + \frac{1}{d}\underbrace{(v_{\rvx}(\mathbf{1})-v_{\rvx}(\mathbf{0})-\mathbf{1}^\top\phi_{\text{fast}}(\rvx;\theta))}_\text{efficiency gap}.
$$
This applies the additive efficiency normalization to the model's outputs for the efficiency constraint.

\subsection{Explanations Analysis}
To evaluate the quality of the ECG explanations, we trained another deep learning model that inputs the same demographics features as the ECG model and features from regions of the ECG that were identified by the algorithm to have high correlation with HbA1c \citep{puli2024explanationsrevealdefinitionencoding}. To do so, we inspected the beat average Shapley values and chose regions of the beat where the absolute Pearson correlation is the highest (reddest/bluest regions in figure). The computed features include:  Average R-peak, Average S-peak, Average RS-interval, Average T-peak, Average T-Onset, and Average Heart Rate and were inputted as tabular features. Compared to the model using only the demographics features (AUROC 0.72 [95\%CI, 0.72-0.74]), the model using such extracted features recovered about 50\% of the performance that the full ECG signal added in addition to the tabular features (AUROC 0.76 [95\%CI, 0.75-0.76]).

\section{Training Details} \label{supp_training}
To train a binary classifer for new-onset diabetes, 
we used HbA1c measured at encounter as the training target.
During training, the value was discretized into four bins: < 5.7\%, 5.7-6.4\%, 6.5-7.9\%, and $\geq$ 8.0\%.
These values were set according to the guidelines \citep{adacare2024}, see \nameref{sec:discussion} for more details on this choice.
We trained the models using an Adam optimizer for 25 epochs to minimize the categorical cross-entropy loss. 
After each epoch, we evaluated the models on the validation set.
We used a learning rate scheduler that multiplies the learning rate by 0.8 after two epochs of no validation loss improvement. 
Early stopping was triggered after the validation loss ceased to improve for five epochs.
The model was trained a single Nvidia A100 GPU.

We used the validation set to select the best network architecture and hyperparameter configuration.
We selected the model with the highest inverse-weighted micro-averaged area under the precision-recall curve (AUPRC).
We examined the effect of varying the number and dimensionality of the fully-connected layers, considering 1, 2, or 3 layers of dimension 100 or 1000.
We considered the effect of temporal dimensionality reduction by modifying the stride lengths to reduce the input 2,500-dimensional vector to either a 10-dimensional vector or an 80-dimensional vector. 
We also tuned the batch size (32, 64, 128) and learning rate ($10^{-5}$, $10^{-4}$, $10^{-3}$). 
Analogously, using the inverse-weighted AUPRC on the validation set, we tuned the number and dimensionality of the fully-connected layers as well as the batch size, and learning rate.

\section{Additional Results} \label{supp:add_results}
\subsection{Additional figures}
Additionally, we calculated model thresholds that correspond to each ADA Risk Test score PPV then assess the TPR for all models. This evaluation quantifies the difference in missed cases. Comparing TPR across models (\figureref{fig:screening_tprs}) matching the PPV of the ADA $\geq 5$, the ECG model again showed significantly better sensitivity at ADA $\geq 5$ (PPV 0.08) (ECG model TPR, 0.94 [95\% CI, 0.92-0.95]; ADA risk test TPR, 0.72 [95\% CI, 0.67-0.77]; QDiabetes-2018 TPR, 0.74 [95\% CI, 0.70-0.77]; p-value $< 0.01$). 
This implies that at the same false alarm rate as the ADA risk test, ECG model was able to capture almost all patients with diabetes, that is, it had far fewer missed cases. 
\input{figures/screening_plots_appendix}

\subsection{Single-Lead ECG Model}
\input{figures/single_lead_auroc}

\subsection{Ablation Studies}

\begin{table}[H]
\centering
\caption{Comparison of AUC across subgroups}
\begin{tabular}{lccc}
\toprule
\textbf{Group (n)} & \textbf{ECG Model} & \textbf{ADA Risk Test} & \textbf{QDiabetes-2018} \\
\midrule
Male (13,902)   & 0.80 [0.79--0.81] & 0.67 [0.66--0.68] & 0.69 [0.67--0.70] \\
Female (15,005) & 0.79 [0.78--0.80] & 0.68 [0.67--0.69] & 0.70 [0.68--0.71] \\
White (18,085)  & 0.81 [0.80--0.82] & 0.71 [0.69--0.72] & 0.69 [0.68--0.70] \\
Black (3,682)   & 0.75 [0.73--0.77] & 0.64 [0.61--0.66] & 0.65 [0.62--0.67] \\
Asian (7,099)   & 0.79 [0.78--0.80] & 0.69 [0.68--0.71] & 0.69 [0.67--0.71] \\
18--40 (4,505)  & 0.83 [0.81--0.87] & 0.74 [0.71--0.78] & 0.72 [0.68--0.76] \\
40--64 (14,914) & 0.81 [0.80--0.82] & 0.68 [0.67--0.69] & 0.70 [0.69--0.71] \\
64+ (9,488)     & 0.74 [0.72--0.75] & 0.59 [0.58--0.61] & 0.62 [0.60--0.64] \\
BMI<18.5 (1,502)& 0.80 [0.77--0.84]	& 0.66 [0.62--0.71]	& 0.68 [0.64--0.73]\\
18.5 $\leq$ BMI < 25 (7,383) & 0.85 [0.84--0.87] & 0.72 [0.70--0.75] & 0.75 [0.73--0.77]\\
25 $\leq$ BMI < 30 (10,401) & 0.80 [0.78--0.81] & 0.65 [0.64--0.66] & 0.67 [0.65--0.69]\\
BMI $\geq$ 30 (9,971) &	0.74 [0.73--0.75] &	0.62 [0.61--0.63] &	0.64 [0.62--0.65] \\
\bottomrule
\end{tabular}
\end{table}

\subsection{External Validation} \label{sec:ext_val}
We collected data from patients in the external set 
to evaluate how well the model is able to generalize across populations.
The external validation set came from a hospital that was later merged into the hospital system, and we also ensured that no patient crossover occurred between our main cohort and the external set. 
Similar to the test set, we selected encounters from patients to those without prior history of diabetes and defined them as the external validation set. 
No crossover of patients occurs between the distinct locations used for external validation and the original study population, which ensures true external validation. 
We assessed the AUROC, and TPR and PPV at the same high-risk thresholds as the test set.

External validation set 
consisted of 24,273 encounters from 21,975 patients, of which 3.5\% have an HbA1c $\geq 6.5$. 
The characteristics for the external validation cohort can be found in \supptabref{tab:screening_ev}.
The ECG model, again, achieved the best performance (AUC, 0.81 [95\%CI, 0.80-0.82]), and outperforming the two baselines (ADA risk test AUC, 0.73 [95\%CI 0.71-0.74], QDiabetes AUC, 0.74 [95\%CI 0.73-0.76]). 

Using the thresholds derived from the test set, we compared the PPVs across models on the external validation set. 
We again focus on the high-risk threshold, the ECG model showed significantly superior precision (PPV, 0.09 [95\%CI 0.08-0.09]) over ADA risk Test (PPV, 0.06 [95\%CI 0.06-0.06]; p-value<0.01) and QDiabetes (PPV, 0.06 [95\%CI 0.06-0.07]; p-value<0.01), while the recall level is similar to the main analysis (0.72). 
The results were consistent with the test set. 

\newpage
\section{Patient Characteristics}
\input{tables/data_split_table1}
\input{tables/outpatient_cohort_table1}
\input{tables/screening_ev}

%% file: figures/model_architecture.tex
\begin{figure}[htb]
    \centering
    \includegraphics[width=0.5\textwidth]{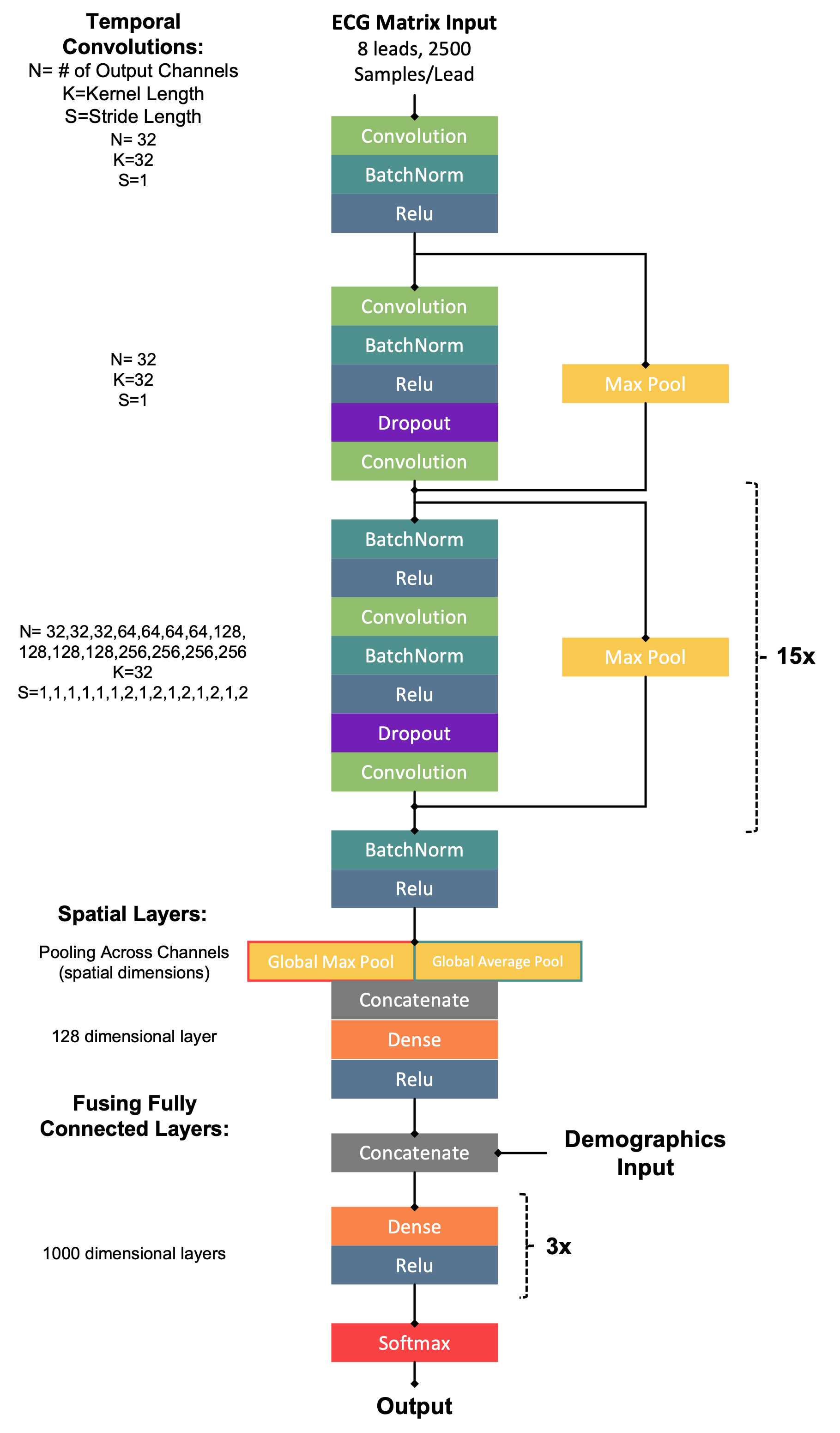}
    \caption{\textbf{ECG model architecture. } A 34-layer ResNet CNN (16 residual connections) processes 8-lead $\times$ 2,500 ECG input through temporal convolutions with batch normalization, ReLU, dropout, and max-pooling. The 128-dimensional ECG embedding is concatenated with demographic features and fed through three fully-connected layers (hidden-size 1000, 1000, 1000) to a softmax output layer predicting four HbA1c bins.}
    \label{fig:model_architecture}
\end{figure}

%% file: tables/covariates.tex
\begin{table}[H]
    \floatconts{tab:variables}{\caption{\textbf{List of Covariates.} For each covariate, the data source is provided. For each model, the set of inputs are identified. The covariate grouping and the corresponding collection period are defined.}}{
    \fontsize{7}{7}\selectfont
    \begin{tabular}{l|c|p{0.1\textwidth}|p{0.1\textwidth}|p{0.1\textwidth}|p{0.1\textwidth}}
        Covariate & Source & ECG Model & ADA Risk Score & QDiabetes-2018 & Propensity Score Models\\ \hline \hline
        \multicolumn{5}{c}{Demographics (During or prior to Encounter)}  \\ \hline
        Age & Epic & \checkmark & \checkmark & \checkmark & \checkmark\\
        Sex & Epic & \checkmark & \checkmark & \checkmark & \checkmark\\
        Race & Epic & \checkmark &  & \checkmark & \checkmark\\
        Ethnicity & Epic & \checkmark &  & \checkmark & \checkmark \\ \hline
        \multicolumn{5}{c}{Vital Signs (During or prior to Encounter)} \\ \hline
        BMI & Epic & \checkmark &  \checkmark& \checkmark& \checkmark \\
        Blood Pressure & Epic & & \checkmark & & \checkmark \\ \hline
        \multicolumn{5}{c}{Personal History (During or prior to Encounter)}  \\ \hline
        Physical Activity & ICD-10 (Z72.3) &  &  \checkmark&& \checkmark \\
        Smoking History & Patient Dim &  & \checkmark && \checkmark \\ \hline
        \multicolumn{5}{c}{Family History}  \\ \hline
        Diabetes & Epic  &  & \checkmark & \checkmark & \checkmark \\
        Cardiovascular Diseases & Epic  &  & & & \checkmark \\ \hline
         \multicolumn{5}{c}{Labs (Prior to Encounter)}  \\ \hline
        HDL & Epic & & &  & \checkmark \\
        LDL & Epic &&  &  & \checkmark \\
        TG & Epic & & &  & \checkmark \\
        HbA1c & Epic & & &  & \checkmark \\ \hline
        \multicolumn{5}{c}{Diagnoses (Prior to Encounter)}  \\ \hline
        Type 1 Diabetes & ICD-10 (E10) &  &&  & \checkmark \\ 
        Type 2 Diabetes & ICD-10 (E11)&  &&  & \checkmark \\ 
        Gestational diabetes & ICD-10 (O24.4)&  &&  & \checkmark \\ 
        PCOS & ICD-10 (E28.2)&  &  && \checkmark \\ 
        Atherosclerosis & ICD-10 (I70,I75)&&  &  & \checkmark \\ 
        Ischemic Heart disease & ICD-10 (I20-26)&  &&  & \checkmark \\ 
        Heart Failure & ICD-10 (I50)&  &&  & \checkmark \\ 
        Cerebrovascular disease & ICD-10 (I60-69)&  &&  & \checkmark \\ 
        Peripheral vascular disease & ICD-10 (I73)& & &  & \checkmark \\ 
        Arrhythmia & ICD-10 (I48,I49) &  &&  & \checkmark \\
        Hypertension & ICD-10 (I10-16)&  & \checkmark && \checkmark \\ 
        Hypercholesterolemia & ICD-10 (E78.0) & & &  & \checkmark \\
        Hyperlipidemia & ICD-10 (E78.2-78.6) &  &&  & \checkmark \\
        Hyperglycemia & ICD-10 (E78.1,R73) &  & & & \checkmark \\ \hline
        \multicolumn{5}{c}{Diabetic Complications (Prior to Encounter)}  \\ \hline
        Retinopathy & ICD-10 (E11.3)& & &  &\checkmark \\ 
        Neuropathy & ICD-10 (E11.4)&  &&  &\checkmark \\ 
        Nephropathy & ICD-10 (E11.2)&  &&  &  \checkmark \\ 
        Other diabetic complications & ICD-10 (E13)&  &&  & \checkmark \\  \hline
        \multicolumn{5}{c}{Acute Conditions (During Encounter)}  \\ \hline
        Polydipsia  & ICD-10 (R63.1)& & &  & \checkmark \\ 
        Polyuria & ICD-10 (R35)&  &  && \checkmark \\ 
        Polyphagia  & ICD-10 (R63.2)&&  &  & \checkmark \\ 
        Weight Loss  & ICD-10 (R63.4)&&  &  & \checkmark \\ 
        Chest pain & ICD-10 (R07.1,R07.8-9)& & &  & \checkmark \\ 
        SOB  & ICD-10 (R06)&  & & & \checkmark \\ 
        Dizziness  & ICD-10 (R42)&  & & & \checkmark \\ \hline
        \multicolumn{5}{c}{Diabetic Medications (Currently Prescribed)}  \\ \hline
        Insulins & Epic&&&&\checkmark \\
        Amylinomimetic & Epic&&&&\checkmark \\
        Biguanides & Epic&&&&\checkmark \\
        Alpha-glucosides inhibitors & Epic&&&&\checkmark \\
        DPP-4 inhibitors & Epic&&&&\checkmark \\
        GLP-1 receptor agonists & Epic&&&&\checkmark \\
        Meglitinides  & Epic&&&&\checkmark \\
        SGLT 2 inhibitors & Epic&&&&\checkmark \\
        Sulfonylureas & Epic&&&&\checkmark \\
        Thiazolidinediones & Epic&&&&\checkmark \\ \hline
        \multicolumn{5}{c}{Cardiovascular Medications (Currently Prescribed)}  \\ \hline
        ACE inhibitors & Epic&&&&\checkmark \\
        A-II Receptor Blockers & Epic&&&&\checkmark \\
        Beta blockers & Epic&&&&\checkmark \\
        Cholesterol lowering & Epic&&&&\checkmark \\
        Calcium Channel Blockers & Epic&&&&\checkmark \\
        Diuretics & Epic&&&&\checkmark \\
        Vasodilators & Epic&&&&\checkmark \\
        Digitalis Preparations & Epic&&&&\checkmark \\
        Antiplatelet & Epic&&&&\checkmark \\
        Anticoagulants  & Epic&&&&\checkmark \\
        Antihypertensives & Epic&&&\checkmark& \\
        \multicolumn{5}{c}{Other Medications (Currently Prescribed)}  \\ \hline
        Corticosteroids & Epic&&&\checkmark& \\
    \end{tabular}
    }
\end{table}

%% file: figures/calibration_curve.tex
\begin{figure}[h]
    \floatconts{fig:calibration_curve}{\caption{\textbf{Calibration curve for the probability of measurement model on held-out test data.}  The probability estimates outputted by the model plotted against the true frequency of the positive label for binned probabilities. Plotted below the curve is a histogram depicting the number of encounters included in each bin. The low calibration error and similarity to a perfectly calibrated curve indicate that the model is well-calibrated. Further, bins with small calibration errors tend to have more samples. Results showed the measurement probability model is well-calibrated.} }{
    \includegraphics[width=0.35\textwidth]{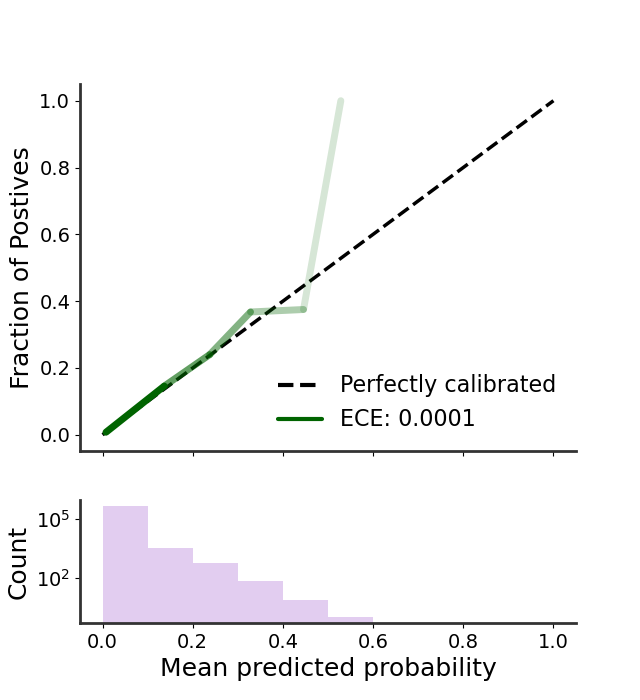}
    }
\end{figure}

%% file: figures/additional_sensitivity.tex
\begin{figure*}[htb]
    \floatconts{fig:screening_sensitivity}{\caption{\textbf{Additional sensitivity analysis.} The ECG model is robust to inaccuracies in estimating the probability of HbA1c acquisition, requiring relatively large re-weighting schemes to invalidate the superiority of the ECG model in terms of the F1-score (left) and PPV (right). The sensitivity weight determines the multiple by which the inverse probability weights can be altered. The ECG model’s performance (F1-score) is selectively minimized as the sensitivity weight increases. The effect of extreme inaccuracies, omitting the contribution of age or the inverse weight entirely (unweighted), is provided as context to indicate that an invalidating re-weighting scheme would be implausible.}}{ 
    \includegraphics[width=.88\textwidth]{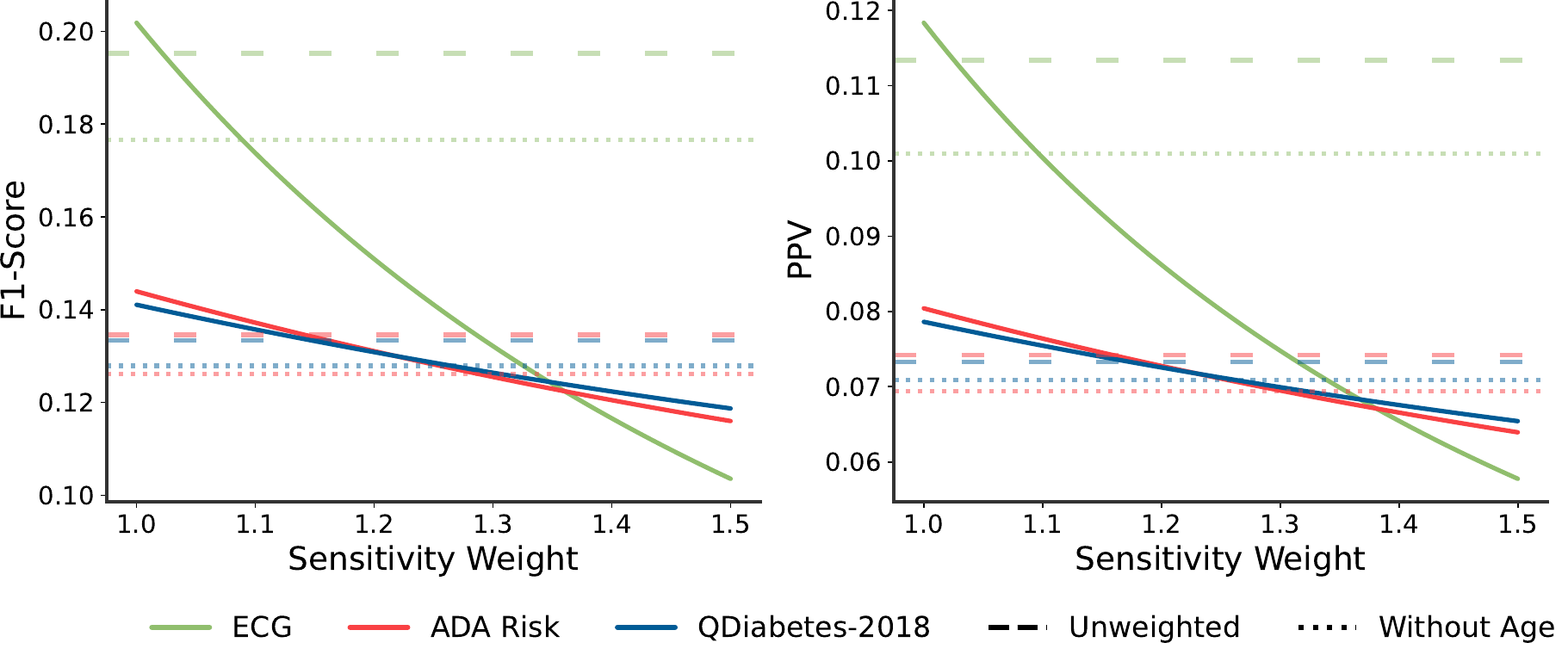}
    }
\end{figure*}

%% file: figures/screening_plots_appendix.tex
\begin{figure*}[htb]  %
  \floatconts{fig:screening_appendix}{\caption{\small{\textbf{New-onset diabetes assessment results.} \textbf{(a)} The ECG model is more discriminative of new-onset diabetes as indicated by the IPW receiver operator curve (ROC). \textbf{(b)} The ECG model has a higher positive predictive value (PPV) for detecting new-onset diabetes at the true positive rate (TPR) corresponding to the ADA's Risk Test threshold of $\geq$5. \textbf{(c)} At ADA $\geq 5$, the ECG model also had superior TPR than ADA Risk Test and QDiabetes-2018. }}}{
    \subfigure[Receiver operator curve]{\label{fig:screening_roc}\includegraphics[width=.30\textwidth]{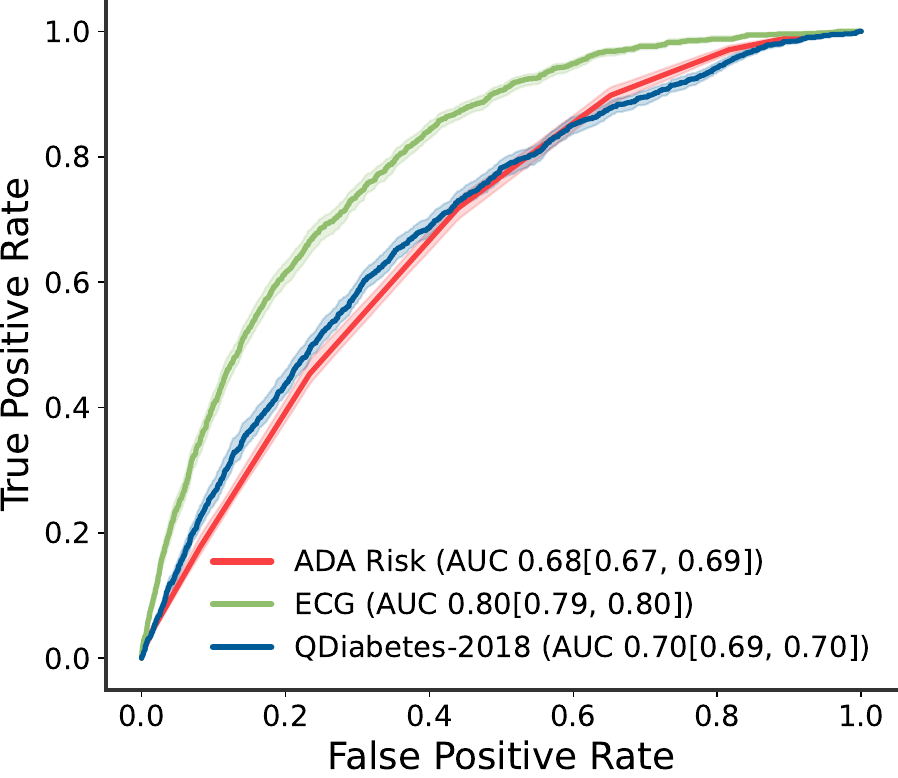}}
   \subfigure[Positive predictive values at each threshold]{\label{fig:screening_ppvs}\includegraphics[width=.30\textwidth]{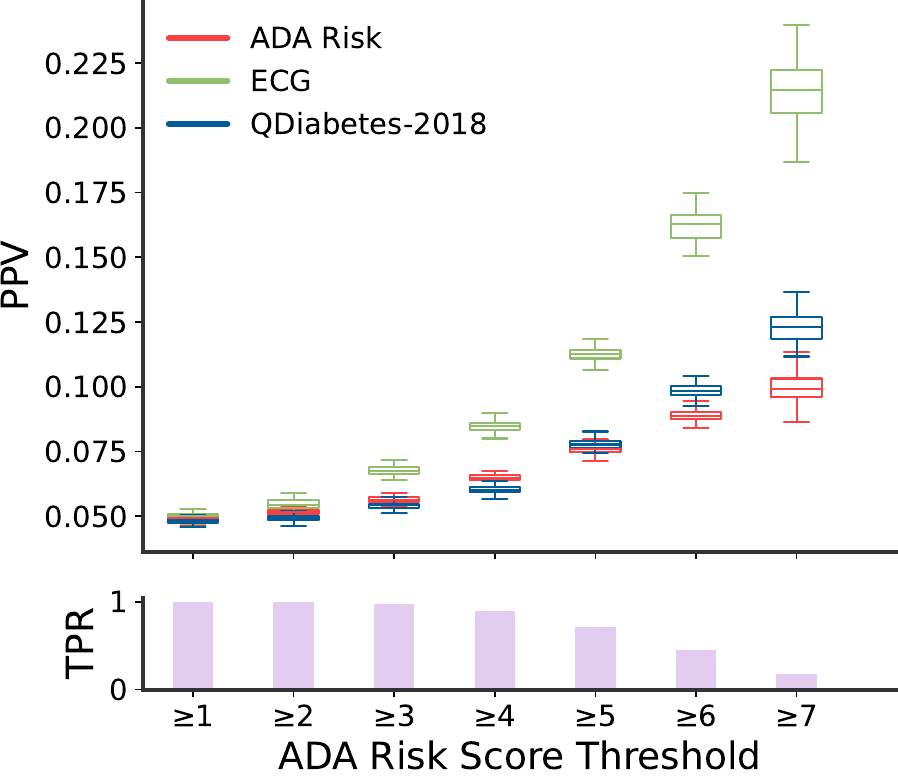}}
   \subfigure[True positive rate at each threshold]{\label{fig:screening_tprs}\includegraphics[width=.28\textwidth]{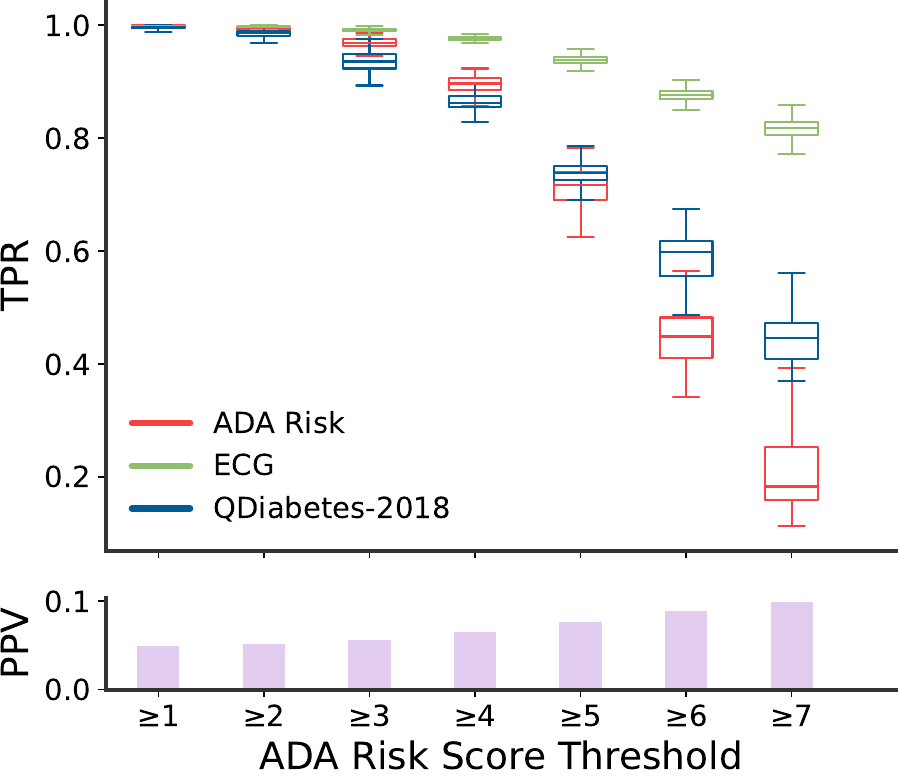}}}
\end{figure*}

%% file: figures/single_lead_auroc.tex
\begin{figure}[htb]
    \centering
    \includegraphics[width=0.4\textwidth]{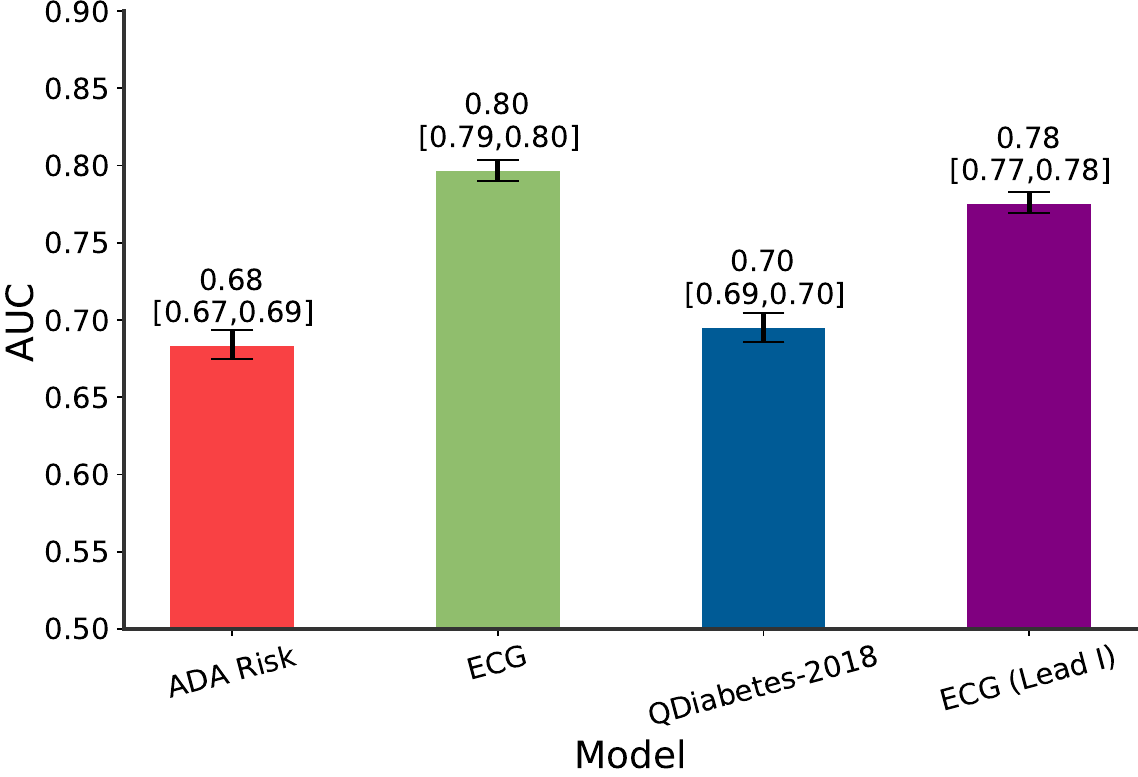}
    \caption{\textbf{Area under the receiver operator curve (AUC) for a model using single lead ECGs.} The AUCs are plotted for each method, with the results for a model using single-lead ECGs plotted in purple. These results indicate that the single-lead ECG model outperforms both the ADA Risk Test and QDiabetes-2018; however, it under-performs the 12-Lead ECG model when estimating the likelihood of new-onset diabetes.}
    \label{fig:single_lead_auroc}
\end{figure}

%% file: tables/data_split_table1.tex
\begin{table*}[htbp]
    \floatconts{tab:data_split_table1}{\caption{Patient Characteristics for the train, validation, and test set. The test contains patients without history of diabetes prior to the visit.}}{
    \fontsize{7}{7.7}\selectfont
    \begin{tabular}{llllll}
        \hline
                                                     & Missing & Overall             & Train               & Validation          & Test             \\
        \hline
        n                                            &         & 199883              & 128209              & 42767               & 28907               \\
        Age, median [Q1,Q3]                          & 357     & 60.7 [49.5,70.9]    & 61.3 [50.2,71.4]    & 61.2 [50.2,71.4]    & 57.2 [45.8,67.6]    \\
        Sex, n (\%) \\
        \qquad  Female                               &         & 108215 (50.6)       & 64948 (50.7)        & 21772 (50.9)        & 15180 (51.9)        \\
        Smoking, n (\%)  \\
        \qquad  Never                                & 3301     & 121485 (61.8)       & 77327 (61.3)        & 26032 (61.9)        & 18126 (64.0)       \\
        \qquad  Former                               &          & 61205 (31.1)        & 39969 (31.7)        & 13199 (31.4)        & 8037 (28.4)        \\
        \qquad Current                               &          & 13892 (7.1)         & 8897 (7.1)          & 2852 (6.8)          & 2143 (7.6)         \\
        Race, n (\%) \\
        \qquad  White                                & 42700   & 123127 (78.3)       & 78721 (78.1)        & 26321 (78.2)        & 18085 (79.7)        \\
        \qquad  Black                                &         & 27403 (17.4)        & 17834 (17.7)        & 5887 (17.5)         & 3682 (16.2)         \\
        \qquad  Asian                                &         & 5626 (3.6)          & 3591 (3.6)          & 1243 (3.7)          & 792 (3.5)           \\
        \qquad  Multiracial                          &         & 635 (0.4)           & 418 (0.4)           & 117 (0.3)           & 100 (0.4)           \\
        \qquad  Pacific Islander                     &         & 392 (0.2)           & 273 (0.3)           & 78 (0.2)            & 41 (0.2)            \\
        Ethnicity, n (\%) \\
        \qquad  Hispanic                             &         & 4191 (9.3)          & 2743 (9.3)          & 949 (9.6)           & 499 (8.8)           \\
        BMI, median [Q1,Q3]                          & 16047   & 28.5 [25.0,33.0]    & 28.6 [25.1,33.1]    & 28.7 [25.1,33.2]    & 28.0 [24.6,32.2]    \\
        Systolic BP, median [Q1,Q3]                  & 2992    & 124.0 [115.0,136.0] & 124.0 [116.0,137.0] & 124.0 [115.0,137.0] & 122.0 [113.0,134.0] \\
        Diastolic BP, median [Q1,Q3]                 & 2992    & 77.0 [70.0,81.0]    & 77.0 [70.0,81.0]    & 77.0 [70.0,81.0]    & 78.0 [70.0,82.0]    \\
        HbA1c, median [Q1,Q3]                        &        & 5.7 [5.3,6.2]       & 5.7 [5.4,6.3]       & 5.7 [5.4,6.3]       & 5.5 [5.3,5.8]      \\
        Prior HbA1c, median [Q1,Q3]                  & 67541   & 5.7 [5.4,6.3]       & 5.8 [5.4,6.5]       & 5.8 [5.4,6.4]       & 5.5 [5.3,5.8]       \\
        Total cholesterol, median [Q1,Q3]            & 60683   & 176.0 [149.0,205.0] & 175.0 [147.0,204.0] & 175.0 [147.0,204.0] & 185.0 [159.0,211.0] \\
        HDL, median [Q1,Q3]                          & 106243  & 52.0 [42.0,63.0]    & 51.0 [42.0,63.0]    & 51.0 [42.0,63.0]    & 54.0 [44.0,66.0]    \\
        LDL, median [Q1,Q3]                          & 90810   & 95.0 [72.0,120.0]   & 93.0 [70.0,119.0]   & 93.0 [70.0,119.0]   & 103.0 [81.0,126.0]  \\
        TG, median [Q1,Q3]                           & 83079   & 107.0 [76.0,154.0]  & 108.0 [77.0,157.0]  & 109.0 [77.0,155.0]  & 100.0 [72.0,142.0]  \\
        eGFR \ensuremath{>} 60 mL/min/1.73m2, n (\%) & 70114   & 112302 (86.5)       & 72427 (85.6)        & 24289 (85.9)        & 15586 (92.2)        \\
        Type 1 DM, n (\%)                            &         & 5162 (2.6)          & 3979 (3.1)          & 1183 (2.8)          &                     \\
        Type 2 DM, n (\%)                            &         & 53273 (26.7)        & 40054 (31.2)        & 13219 (30.9)        &                     \\
        Hyperglycemia, n (\%)                        &         & 66318 (33.2)        & 42834 (33.4)        & 14167 (33.1)        & 9317 (32.2)         \\
        FHx cardiovascular disease, n (\%)           &         & 97042 (48.5)        & 62602 (48.8)        & 20820 (48.7)        & 13620 (47.1)        \\
        FHx Diabetes, n (\%)                         &         & 70147 (35.1)        & 45973 (35.9)        & 15358 (35.9)        & 8816 (30.5)         \\
        Gestational Diabetes, n (\%)                 &         & 551 (0.3)           & 355 (0.3)           & 112 (0.3)           & 84 (0.3)            \\
        Polycystic ovarian syndrome, n (\%)          &         & 1346 (0.7)          & 856 (0.7)           & 293 (0.7)           & 197 (0.7)           \\
        Poor diet, n (\%)                            &         & 43 (0.0)            & 22 (0.0)            & 15 (0.0)            & 6 (0.0)             \\
        Lack of physical activity, n (\%)            &         & 108 (0.1)           & 74 (0.1)            & 19 (0.0)            & 15 (0.1)            \\
        Hypertension, n (\%)                         &         & 104484 (52.3)       & 69987 (54.6)        & 23149 (54.1)        & 11348 (39.3)        \\
        Hypercholesterolemia, n (\%)                 &         & 52518 (26.3)        & 34924 (27.2)        & 11784 (27.6)        & 5810 (20.1)         \\
        Hyperlipidemia, n (\%)                       &         & 103899 (52.0)       & 69018 (53.8)        & 22980 (53.7)        & 11901 (41.2)        \\
        Cardiovascular disease, n (\%)               &         & 12945 (6.5)         & 8734 (6.8)          & 2938 (6.9)          & 1273 (4.4)          \\
        Heart failure, n (\%)                        &         & 14360 (7.2)         & 10086 (7.9)         & 3295 (7.7)          & 979 (3.4)           \\
        Ischemic heart disease, n (\%)               &         & 48873 (24.5)        & 33391 (26.0)        & 11039 (25.8)        & 4443 (15.4)         \\
        Peripheral vascular disease, n (\%)          &         & 15382 (7.7)         & 10742 (8.4)         & 3610 (8.4)          & 1030 (3.6)          \\
        Arrhythmia, n (\%)                           &         & 32374 (16.2)        & 21913 (17.1)        & 7136 (16.7)         & 3325 (11.5)         \\
        Atherosclerosis, n (\%)                      &         & 15038 (7.5)         & 10328 (8.1)         & 3524 (8.2)          & 1186 (4.1)          \\
        Cerebrovascular disease, n (\%)              &         & 22506 (11.3)        & 15397 (12.0)        & 4993 (11.7)         & 2116 (7.3)          \\
        Polydipsia, n (\%)                           &         & 74 (0.0)            & 49 (0.0)            & 12 (0.0)            & 13 (0.0)            \\
        Polyphagia, n (\%)                           &         & 59 (0.0)            & 39 (0.0)            & 11 (0.0)            & 9 (0.0)             \\
        Polyuria, n (\%)                             &         & 5699 (2.9)          & 3683 (2.9)          & 1171 (2.7)          & 845 (2.9)           \\
        Weight loss, n (\%)                          &         & 1243 (0.6)          & 794 (0.6)           & 288 (0.7)           & 161 (0.6)           \\
        Chest pain, n (\%)                           &         & 8763 (4.4)          & 5521 (4.3)          & 1847 (4.3)          & 1395 (4.8)          \\
        Dyspnea, n (\%)                              &         & 10922 (5.5)         & 7007 (5.5)          & 2325 (5.4)          & 1590 (5.5)          \\
        Dizziness, n (\%)                            &         & 4351 (2.2)          & 2796 (2.2)          & 922 (2.2)           & 633 (2.2)           \\
        Diabetic retinopathy, n (\%)                 &         & 4738 (2.4)          & 3617 (2.8)          & 1121 (2.6)          &                     \\
        Diabetic nephropathy, n (\%)                 &         & 8114 (4.1)          & 6114 (4.8)          & 2000 (4.7)          &                     \\
        Diabetic neuropathy, n (\%)                  &         & 9028 (4.5)          & 6771 (5.3)          & 2257 (5.3)          &                     \\
        Other diabetic complications, n (\%)         &         & 5908 (3.0)          & 4489 (3.5)          & 1419 (3.3)          &                     \\
        \hline
    \end{tabular}
    }
\end{table*}

%% file: tables/outpatient_cohort_table1.tex
\begin{table*}[!htbp]
    \floatconts{tab:outpatient_table1}{\caption{Patient Characteristics for cohort used to build the probability of HbA1c acquisition model stratified by whether an HbA1c were measured given that ECG was measured.}}{
    \fontsize{7}{7.7}\selectfont
\begin{tabular}{llllll}
\hline
                                                       & Missing   & Overall             & Missing HbA1c   & HbA1c Measured   & P-Value   \\
\hline
 n                                                     &           & 483289              & 414089              & 69200               &           \\
 Age, median [Q1,Q3]                                   & 8386      & 60.3 [47.7,71.6]    & 61.1 [48.4,72.4]    & 56.0 [44.6,66.4]    & \ensuremath{<}0.001    \\
 Sex, n (\%)   \\
     \qquad Female           & 0         & 255885 (52.9)       & 220159 (53.2)       & 35726 (51.6)        & \ensuremath{<}0.001    \\
 Smoking, n (\%) \\
     \qquad Never            & 19459     & 284173 (61.3)       & 239837 (60.5)       & 44336 (65.7)        & \ensuremath{<}0.001    \\
     \qquad Former           &           & 144842 (31.2)       & 127233 (32.1)       & 17609 (26.1)        &           \\
     \qquad Current          &           & 34815 (7.5)         & 29306 (7.4)         & 5509 (8.2)          &           \\
 Race, n (\%)  \\
     \qquad White            & 106352    & 308623 (81.9)       & 269411 (83.1)       & 39212 (74.3)        & \ensuremath{<}0.001    \\
     \qquad Black            &           & 53557 (14.2)        & 42571 (13.1)        & 10986 (20.8)        &           \\
     \qquad Asian            &           & 12710 (3.4)         & 10508 (3.2)         & 2202 (4.2)          &           \\
     \qquad Pacific Islander &           & 883 (0.2)           & 782 (0.2)           & 101 (0.2)           &           \\
     \qquad Multiracial      &           & 1164 (0.3)          & 924 (0.3)           & 240 (0.5)           &           \\
 Ethnicity, n (\%)   \\
      \qquad Hispanic         &           & 9313 (9.6)          & 8203 (9.5)          & 1110 (10.2)         &           \\
BMI, median [Q1,Q3]                 & 61450     & 27.5 [24.1,31.9]    & 27.5 [24.0,31.7]    & 28.2 [24.7,32.7]    & \ensuremath{<}0.001    \\
 Systolic BP, median [Q1,Q3]         & 20491     & 124.0 [114.0,136.0] & 124.0 [114.0,136.0] & 124.0 [114.0,136.0] & 0.167     \\
 Diastolic BP, median [Q1,Q3]        & 20491     & 76.0 [70.0,80.0]    & 75.0 [70.0,80.0]    & 78.0 [70.0,82.0]    & \ensuremath{<}0.001    \\
 HbA1c, median [Q1,Q3]               & 407616    & 5.6 [5.3,6.0]       & 5.6 [5.3,6.1]       & 5.6 [5.3,6.0]       & \ensuremath{<}0.001    \\
 Prior HbA1c, median [Q1,Q3]         & 300369    & 5.6 [5.3,6.1]       & 5.6 [5.3,6.1]       & 5.6 [5.3,6.1]       & 0.623     \\
 Total cholesterol, median [Q1,Q3]   & 250681    & 180.0 [152.0,208.0] & 178.0 [151.0,207.0] & 185.0 [159.0,212.0] & \ensuremath{<}0.001    \\
 HDL, median [Q1,Q3]                 & 338804    & 53.0 [43.0,65.0]    & 52.0 [43.0,65.0]    & 53.0 [44.0,65.0]    & \ensuremath{<}0.001    \\
 LDL, median [Q1,Q3]                 & 306155    & 98.0 [75.0,122.0]   & 96.0 [74.0,121.0]   & 103.0 [80.0,126.0]  & \ensuremath{<}0.001    \\
 TG, median [Q1,Q3]                  & 293032    & 103.0 [73.0,148.0]  & 102.0 [73.0,148.0]  & 105.0 [74.0,151.0]  & \ensuremath{<}0.001    \\
Type 1 DM, n (\%)                    &         & 5742 (1.2)          & 4853 (1.2)          & 889 (1.3)           & 0.012     \\
 Type 2 DM, n (\%)                    &         & 74711 (15.5)        & 62303 (15.0)        & 12408 (17.9)        & \ensuremath{<}0.001    \\
 Hyperglycemia, n (\%)                &         & 82895 (17.2)        & 66889 (16.2)        & 16006 (23.1)        & \ensuremath{<}0.001    \\
 FHx cardiovascular disease, n (\%)   &         & 202350 (41.9)       & 173667 (41.9)       & 28683 (41.4)        & 0.016     \\
 FHx Diabetes, n (\%)                 &         & 127541 (26.4)       & 104386 (25.2)       & 23155 (33.5)        & \ensuremath{<}0.001    \\
 Gestational Diabetes, n (\%)         &         & 1157 (0.2)          & 959 (0.2)           & 198 (0.3)           & 0.007     \\
 Polycystic ovarian syndrome, n (\%)  &         & 2953 (0.6)          & 2463 (0.6)          & 490 (0.7)           & \ensuremath{<}0.001    \\
 Poor diet, n (\%)                    &         & 92 (0.0)            & 78 (0.0)            & 14 (0.0)            & 0.922     \\
 Lack of physical activity, n (\%)    &         & 168 (0.0)           & 143 (0.0)           & 25 (0.0)            & 0.922     \\
 Hypertension, n (\%)                 &         & 212319 (43.9)       & 186566 (45.1)       & 25753 (37.2)        & \ensuremath{<}0.001    \\
 Hypercholesterolemia, n (\%)         &         & 86854 (18.0)        & 74995 (18.1)        & 11859 (17.1)        & \ensuremath{<}0.001    \\
 Hyperlipidemia, n (\%)               &         & 189432 (39.2)       & 164079 (39.6)       & 25353 (36.6)        & \ensuremath{<}0.001    \\
 Cardiovascular disease, n (\%)       &         & 25993 (5.4)         & 24235 (5.9)         & 1758 (2.5)          & \ensuremath{<}0.001    \\
 Heart failure, n (\%)                &         & 30443 (6.3)         & 28960 (7.0)         & 1483 (2.1)          & \ensuremath{<}0.001    \\
 Ischemic heart disease, n (\%)       &         & 95398 (19.7)        & 88051 (21.3)        & 7347 (10.6)         & \ensuremath{<}0.001    \\
 Peripheral vascular disease, n (\%)  &         & 24330 (5.0)         & 21866 (5.3)         & 2464 (3.6)          & \ensuremath{<}0.001    \\
 Arrhythmia, n (\%)                   &         & 76555 (15.8)        & 71941 (17.4)        & 4614 (6.7)          & \ensuremath{<}0.001    \\
 Atherosclerosis, n (\%)              &         & 25960 (5.4)         & 23857 (5.8)         & 2103 (3.0)          & \ensuremath{<}0.001    \\
 Cerebrovascular disease, n (\%)      &         & 41829 (8.7)         & 37991 (9.2)         & 3838 (5.5)          & \ensuremath{<}0.001    \\
 Polydipsia, n (\%)                   &         & 52 (0.0)            & 27 (0.0)            & 25 (0.0)            & \ensuremath{<}0.001    \\
 Polyphagia, n (\%)                   &         & 171 (0.0)           & 159 (0.0)           & 12 (0.0)            & 0.009     \\
 Polyuria, n (\%)                     &         & 5370 (1.1)          & 3480 (0.8)          & 1890 (2.7)          & \ensuremath{<}0.001    \\
 Weight loss, n (\%)                  &         & 1820 (0.4)          & 1421 (0.3)          & 399 (0.6)           & \ensuremath{<}0.001    \\
 Chest pain, n (\%)                   &         & 23422 (4.8)         & 19840 (4.8)         & 3582 (5.2)          & \ensuremath{<}0.001    \\
 Dyspnea, n (\%)                      &         & 27597 (5.7)         & 23993 (5.8)         & 3604 (5.2)          & \ensuremath{<}0.001    \\
 Dizziness, n (\%)                    &         & 8789 (1.8)          & 7202 (1.7)          & 1587 (2.3)          & \ensuremath{<}0.001    \\
 Diabetic retinopathy, n (\%)         &         & 4784 (1.0)          & 4066 (1.0)          & 718 (1.0)           & 0.178     \\
 Diabetic nephropathy, n (\%)         &         & 8734 (1.8)          & 7811 (1.9)          & 923 (1.3)           & \ensuremath{<}0.001    \\
 Diabetic neuropathy, n (\%)          &         & 9503 (2.0)          & 8086 (2.0)          & 1417 (2.0)          & 0.099     \\
 Other diabetic complications, n (\%) &         & 5996 (1.2)          & 5017 (1.2)          & 979 (1.4)           & \ensuremath{<}0.001    \\
\hline
\end{tabular}
}
\end{table*}


%% file: tables/screening_ev.tex
\begin{table*}[!htbp]
    \floatconts{tab:screening_ev}{\caption{Patient Characteristics for the external set stratified by whether the patient is newly diagnosed with diabetes (HbA1c >6.4).}}{
    \fontsize{7}{7.7}\selectfont 
\begin{tabular}{lllll}
\hline
                                                      & Missing   & Overall             & Diabetic            & Normal                  \\
\hline
 n                                                    &           & 24273               & 840                 & 23433                          \\
 Age, median [Q1,Q3]                                  & 29        & 57.2 [44.2,68.4]    & 63.3 [53.9,72.5]    & 56.9 [43.7,68.2]        \\
 Sex, n (\%)   \\
 \qquad Female           &           & 12583 (51.8)        & 338 (40.2)          & 12245 (52.3)            \\
 Smoking, n (\%)  \\
 \qquad Never            & 421       & 15634 (65.5)        & 497 (60.9)          & 15137 (65.7)              \\
 \qquad Former           &           & 6565 (27.5)         & 260 (31.9)          & 6305 (27.4)                    \\
 \qquad Current          &           & 1653 (6.9)          & 59 (7.2)            & 1594 (6.9)                     \\
 Race, n (\%)   \\
 \qquad White            & 5664      & 16246 (87.2)        & 491 (77.6)          & 15755 (87.5)            \\
 \qquad Black            &           & 1801 (9.7)          & 111 (17.5)          & 1690 (9.4)                     \\
 \qquad Asian            &           & 452 (2.4)           & 24 (3.8)            & 428 (2.4)                      \\
 \qquad Multiracial      &           & 107 (0.6)           & 5 (0.8)             & 102 (0.6)                      \\
 \qquad Pacific Islander &           & 23 (0.1)            & 2 (0.3)             & 21 (0.1)                       \\
 Ethnicity, n (\%)    \\
 \qquad Hispanic         &           & 79 (3.7)            & 2 (2.8)             & 77 (3.7)            \\
 BMI, median [Q1,Q3]                                  & 945       & 27.3 [24.1,31.2]    & 31.6 [27.5,36.2]    & 27.2 [24.0,31.1]    \\
 Systolic BP, median [Q1,Q3]                          & 130       & 122.0 [114.0,130.0] & 128.5 [120.0,140.0] & 121.0 [114.0,130.0] \\
 Diastolic BP, median [Q1,Q3]                         & 130       & 75.0 [70.0,80.0]    & 80.0 [70.0,82.0]    & 74.0 [70.0,80.0]    \\
 HbA1c, median [Q1,Q3]                                & 0         & 5.5 [5.2,5.8]       & 6.9 [6.6,7.6]       & 5.5 [5.2,5.7]       \\
 Prior HbA1c, median [Q1,Q3]                          & 7101      & 5.5 [5.2,5.7]       & 6.5 [6.2,6.7]       & 5.5 [5.2,5.7]       \\
 Total cholesterol, median [Q1,Q3]                    & 5860      & 184.0 [159.0,210.0] & 179.0 [152.0,205.5] & 184.0 [160.0,210.0] \\
 HDL, median [Q1,Q3]                                  & 10152     & 56.0 [46.0,68.0]    & 47.0 [38.2,55.0]    & 56.0 [46.0,68.0]    \\
 LDL, median [Q1,Q3]                                  & 9647      & 101.0 [80.0,124.0]  & 99.5 [75.0,120.0]   & 101.0 [80.0,124.0]  \\
 TG, median [Q1,Q3]                                   & 10792     & 105.0 [75.0,148.0]  & 141.0 [107.0,208.0] & 104.0 [74.0,147.0]  \\
 eGFR \ensuremath{>} 60 mL/min/1.73m2, n (\%)                      & 9512      & 13607 (92.2)        & 371 (89.0)          & 13236 (92.3)        \\
 Hyperglycemia, n (\%)                                &           & 8836 (36.4)         & 359 (42.7)          & 8477 (36.2)         \\
 FHx cardiovascular disease, n (\%)                   &           & 9375 (38.6)         & 362 (43.1)          & 9013 (38.5)         \\
 FHx Diabetes, n (\%)                                 &           & 5448 (22.4)         & 282 (33.6)          & 5166 (22.0)         \\
 Gestational Diabetes, n (\%)                         &           & 90 (0.4)            & 5 (0.6)             & 85 (0.4)            \\
 Polycystic ovarian syndrome, n (\%)                  &           & 136 (0.6)           & 2 (0.2)             & 134 (0.6)           \\
 Poor diet, n (\%)                                    &           & 7 (0.0)             &                     & 7 (0.0)             \\
 Lack of physical activity, n (\%)                    &           & 1 (0.0)             &                     & 1 (0.0)             \\
 Hypertension, n (\%)                                 &           & 9928 (40.9)         & 445 (53.0)          & 9483 (40.5)         \\
 Hypercholesterolemia, n (\%)                         &           & 6011 (24.8)         & 182 (21.7)          & 5829 (24.9)         \\
 Hyperlipidemia, n (\%)                               &           & 11474 (47.3)        & 388 (46.2)          & 11086 (47.3)        \\
 Cardiovascular disease, n (\%)                       &           & 838 (3.5)           & 30 (3.6)            & 808 (3.4)           \\
 Heart failure, n (\%)                                &           & 813 (3.3)           & 54 (6.4)            & 759 (3.2)           \\
 Ischemic heart disease, n (\%)                       &           & 2951 (12.2)         & 160 (19.0)          & 2791 (11.9)         \\
 Peripheral vascular disease, n (\%)                  &           & 870 (3.6)           & 37 (4.4)            & 833 (3.6)           \\
 Arrhythmia, n (\%)                                   &           & 2713 (11.2)         & 105 (12.5)          & 2608 (11.1)         \\
 Atherosclerosis, n (\%)                              &           & 770 (3.2)           & 24 (2.9)            & 746 (3.2)           \\
 Cerebrovascular disease, n (\%)                      &           & 1488 (6.1)          & 60 (7.1)            & 1428 (6.1)          \\
 Polydipsia, n (\%)                                   &           & 8 (0.0)             & 1 (0.1)             & 7 (0.0)             \\
 Polyphagia, n (\%)                                   &           & 0 (0.0)             & 0 (0.0)             & 0 (0.0)                  \\
 Polyuria, n (\%)                                     &           & 914 (3.8)           & 37 (4.4)            & 877 (3.7)           \\
 Weight loss, n (\%)                                  &           & 112 (0.5)           & 5 (0.6)             & 107 (0.5)           \\
 Chest pain, n (\%)                                   &           & 1139 (4.7)          & 53 (6.3)            & 1086 (4.6)          \\
 Dyspnea, n (\%)                                      &           & 1785 (7.4)          & 99 (11.8)           & 1686 (7.2)          \\
 Dizziness, n (\%)                                    &           & 426 (1.8)           & 18 (2.1)            & 408 (1.7)           \\
\hline
\end{tabular}
}
\end{table*}

%% file: sample.bib
@article{CDC2020,
  title={National diabetes statistics report},
  author={CDC and others},
  journal={Atlanta, GA: Centers for Disease Control and Prevention, US Department of Health and Human Services},
  year={2024}
}

@inproceedings{jethani2021have,
  title={Have We Learned to Explain?: How Interpretability Methods Can Learn to Encode Predictions in their Interpretations.},
  author={Jethani, Neil and Sudarshan, Mukund and Aphinyanaphongs, Yindalon and Ranganath, Rajesh},
  booktitle={International Conference on Artificial Intelligence and Statistics},
  pages={1459--1467},
  year={2021},
  organization={PMLR}
}

@article{Simmons2017,
author = {Simmons, Rebecca K. and Griffin, Simon J. and Lauritzen, Torsten and Sandb{\ae}k, Annelli},
doi = {10.1007/S00125-017-4299-Y},
issn = {14320428},
journal = {Diabetologia},
month = {nov},
number = {11},
pages = {2192},
pmid = {28831539},
publisher = {Springer},
title = {{Effect of screening for type 2 diabetes on risk of cardiovascular disease and mortality: a controlled trial among 139,075 individuals diagnosed with diabetes in Denmark between 2001 and 2009}},
url = {/pmc/articles/PMC6108415/ /pmc/articles/PMC6108415/?report=abstract https://www.ncbi.nlm.nih.gov/pmc/articles/PMC6108415/},
volume = {60},
year = {2017}
}

@article{Hannun2019,
  title={Cardiologist-level arrhythmia detection and classification in ambulatory electrocardiograms using a deep neural network},
  author={Hannun, Awni Y and Rajpurkar, Pranav and Haghpanahi, Masoumeh and Tison, Geoffrey H and Bourn, Codie and Turakhia, Mintu P and Ng, Andrew Y},
  journal={Nature medicine},
  volume={25},
  number={1},
  pages={65--69},
  year={2019},
  publisher={Nature Publishing Group}
}

@article{Attia2019,
author = {Attia, Zachi I and Friedman, Paul A and Noseworthy, Peter A and Lopez-Jimenez, Francisco and Ladewig, Dorothy J and Satam, Gaurav and Pellikka, Patricia A and Munger, Thomas M and Asirvatham, Samuel J and Scott, Christopher G and Carter, Rickey E and Kapa, Suraj},
doi = {10.1161/CIRCEP.119.007284},
issn = {1941-3084},
journal = {Circulation. Arrhythmia and electrophysiology},
month = {sep},
number = {9},
pages = {e007284},
pmid = {31450977},
title = {{Age and Sex Estimation Using Artificial Intelligence From Standard 12-Lead ECGs.}},
url = {https://www.ahajournals.org/doi/10.1161/CIRCEP.119.007284 http://www.ncbi.nlm.nih.gov/pubmed/31450977},
volume = {12},
year = {2019}
}

@article{Attia2019a,
author = {Attia, Zachi I. and Kapa, Suraj and Lopez-Jimenez, Francisco and McKie, Paul M. and Ladewig, Dorothy J. and Satam, Gaurav and Pellikka, Patricia A. and Enriquez-Sarano, Maurice and Noseworthy, Peter A. and Munger, Thomas M. and Asirvatham, Samuel J. and Scott, Christopher G. and Carter, Rickey E. and Friedman, Paul A.},
doi = {10.1038/s41591-018-0240-2},
issn = {1078-8956},
journal = {Nature Medicine},
month = {jan},
number = {1},
pages = {70--74},
publisher = {Nature Publishing Group},
title = {{Screening for cardiac contractile dysfunction using an artificial intelligence–enabled electrocardiogram}},
url = {http://www.nature.com/articles/s41591-018-0240-2},
volume = {25},
year = {2019}
}

@article{Porumb2020,
author = {Porumb, Mihaela and Stranges, Saverio and Pescap{\`{e}}, Antonio and Pecchia, Leandro},
doi = {10.1038/s41598-019-56927-5},
issn = {2045-2322},
journal = {Scientific Reports},
month = {dec},
number = {1},
pages = {170},
title = {{Precision Medicine and Artificial Intelligence: A Pilot Study on Deep Learning for Hypoglycemic Events Detection based on ECG}},
url = {http://www.nature.com/articles/s41598-019-56927-5},
volume = {10},
year = {2020}
}

@article{Lin2021,
author = {Lin, Chin Sheng and Lee, Yung Tsai and Fang, Wen Hui and Lou, Yu Sheng and Kuo, Feng Chih and Lee, Chia Cheng and Lin, Chin},
doi = {10.3390/JPM11080725},
issn = {20754426},
journal = {Journal of Personalized Medicine},
month = {aug},
number = {8},
pages = {725},
pmid = {34442369},
publisher = {Multidisciplinary Digital Publishing Institute  (MDPI)},
title = {{Deep Learning Algorithm for Management of Diabetes Mellitus via Electrocardiogram-Based Glycated Hemoglobin (ECG-HbA1c): A Retrospective Cohort Study}},
url = {/pmc/articles/PMC8398464/ /pmc/articles/PMC8398464/?report=abstract https://www.ncbi.nlm.nih.gov/pmc/articles/PMC8398464/},
volume = {11},
year = {2021}
}

@article{Zhao,
  title={Sensitivity analysis for inverse probability weighting estimators via the percentile bootstrap},
  author={Zhao, Qingyuan and Small, Dylan S and Bhattacharya, Bhaswar B},
  journal={Journal of the Royal Statistical Society: Series B (Statistical Methodology)},
  volume={81},
  number={4},
  pages={735--761},
  year={2019},
  publisher={Wiley Online Library}
}

@article{Bang2009,
author = {Bang, Heejung and Edwards, Alison M. and Bomback, Andrew S. and Ballantyne, Christie M. and Brillon, David and Callahan, Mark A. and Teutsch, Steven M. and Mushlin, Alvin I. and Kern, Lisa M.},
doi = {10.1059/0003-4819-151-11-200912010-00005},
issn = {00034819},
journal = {Annals of internal medicine},
month = {dec},
number = {11},
pages = {775},
pmid = {19949143},
publisher = {NIH Public Access},
title = {{A patient self-assessment diabetes screening score: development, validation, and comparison to other diabetes risk assessment scores}},
url = {/pmc/articles/PMC3633111/ /pmc/articles/PMC3633111/?report=abstract https://www.ncbi.nlm.nih.gov/pmc/articles/PMC3633111/},
volume = {151},
year = {2009}
}

@article{Hofman2021,
author = {Hofman, Jake M and Watts, Duncan J and Athey, Susan and Garip, Filiz and Griffiths, Thomas L and Kleinberg, Jon and Margetts, Helen and Mullainathan, Sendhil and Salganik, Matthew J and Vazire, Simine and Vespignani, Alessandro and Yarkoni, Tal},
doi = {10.1038/s41586-021-03659-0},
journal = {Nature},
pages = {181},
title = {{Integrating explanation and prediction in computational social science}},
url = {https://doi.org/10.1038/s41586-021-03659-0},
volume = {595},
year = {2021}
}

@article{harris2000early,
  title={Early detection of undiagnosed diabetes mellitus: a US perspective},
  author={Harris, Maureen I and Eastman, Richard C},
  journal={Diabetes/metabolism research and reviews},
  volume={16},
  number={4},
  pages={230--236},
  year={2000},
  publisher={Wiley Online Library}
}

@article{dall2019economic,
  title={The economic burden of elevated blood glucose levels in 2017: diagnosed and undiagnosed diabetes, gestational diabetes mellitus, and prediabetes},
  author={Dall, Timothy M and Yang, Wenya and Gillespie, Karin and Mocarski, Michelle and Byrne, Erin and Cintina, Inna and Beronja, Kaleigh and Semilla, April P and Iacobucci, William and Hogan, Paul F},
  journal={Diabetes care},
  volume={42},
  number={9},
  pages={1661--1668},
  year={2019},
  publisher={Am Diabetes Assoc}
}

@article{mitani2020detection,
  title={Detection of anaemia from retinal fundus images via deep learning},
  author={Mitani, Akinori and Huang, Abigail and Venugopalan, Subhashini and Corrado, Greg S and Peng, Lily and Webster, Dale R and Hammel, Naama and Liu, Yun and Varadarajan, Avinash V},
  journal={Nature Biomedical Engineering},
  volume={4},
  number={1},
  pages={18--27},
  year={2020},
  publisher={Nature Publishing Group}
}

@article{hughes2021deep,
  title={Deep learning evaluation of biomarkers from echocardiogram videos},
  author={Hughes, J Weston and Yuan, Neal and He, Bryan and Ouyang, Jiahong and Ebinger, Joseph and Botting, Patrick and Lee, Jasper and Theurer, John and Tooley, James E and Nieman, Koen and others},
  journal={EBioMedicine},
  volume={73},
  pages={103613},
  year={2021},
  publisher={Elsevier}
}

@article{qdb,
	author = {Hippisley-Cox, Julia and Coupland, Carol},
	doi = {10.1136/bmj.j5019},
	elocation-id = {j5019},
	eprint = {https://www.bmj.com/content/359/bmj.j5019.full.pdf},
	issn = {0959-8138},
	journal = {BMJ},
	publisher = {BMJ Publishing Group Ltd},
	title = {Development and validation of QDiabetes-2018 risk prediction algorithm to estimate future risk of type 2 diabetes: cohort study},
	url = {https://www.bmj.com/content/359/bmj.j5019},
	volume = {359},
	year = {2017}}

@article{tableone,
    author = {Pollard, Tom J and Johnson, Alistair E W and Raffa, Jesse D and Mark, Roger G},
    title = "{tableone: An open source Python package for producing summary statistics for research papers}",
    journal = {JAMIA Open},
    volume = {1},
    number = {1},
    pages = {26-31},
    year = {2018},
    month = {05},
    issn = {2574-2531},
    doi = {10.1093/jamiaopen/ooy012},
    url = {https://doi.org/10.1093/jamiaopen/ooy012},
    eprint = {https://academic.oup.com/jamiaopen/article-pdf/1/1/26/32298238/ooy012.pdf},
}

@article{scikit-learn,
 title={Scikit-learn: Machine Learning in {P}ython},
 author={Pedregosa, F. and Varoquaux, G. and Gramfort, A. and Michel, V.
         and Thirion, B. and Grisel, O. and Blondel, M. and Prettenhofer, P.
         and Weiss, R. and Dubourg, V. and Vanderplas, J. and Passos, A. and
         Cournapeau, D. and Brucher, M. and Perrot, M. and Duchesnay, E.},
 journal={Journal of Machine Learning Research},
 volume={12},
 pages={2825--2830},
 year={2011}
}

@misc{jethani2022fastshap,
      title={FastSHAP: Real-Time Shapley Value Estimation}, 
      author={Neil Jethani and Mukund Sudarshan and Ian Covert and Su-In Lee and Rajesh Ranganath},
      year={2022},
      eprint={2107.07436},
      archivePrefix={arXiv},
      primaryClass={stat.ML}
}

@misc{jethani2023dont,
      title={Don't be fooled: label leakage in explanation methods and the importance of their quantitative evaluation}, 
      author={Neil Jethani and Adriel Saporta and Rajesh Ranganath},
      year={2023},
      eprint={2302.12893},
      archivePrefix={arXiv},
      primaryClass={cs.LG}
}

@article{adacare2024,
	author = {ElSayed, Nuha A. and Aleppo, Grazia and Bannuru, Raveendhara R. and Bruemmer, Dennis and Collins, Billy S. and Ekhlaspour, Laya and Gaglia, Jason L. and Hilliard, Marisa E. and Johnson, Eric L. and Khunti, Kamlesh and Lingvay, Ildiko and Matfin, Glenn and McCoy, Rozalina G. and Perry, Mary Lou and Pilla, Scott J. and Polsky, Sarit and Prahalad, Priya and Pratley, Richard E. and Segal, Alissa R. and Seley, Jane Jeffrie and Selvin, Elizabeth and Stanton, Robert C. and Gabbay, Robert A.},
	journal = {Diabetes Care},
	number = {Supplement 1},
	pages = {S20-S42},
	title = {2. Diagnosis and Classification of Diabetes: Standards of Care in Diabetes 2024},
	volume = {47},
	year = {2023}}

@article{mldm_1,
	author = {Nomura, Akihiro and Noguchi, Masahiro and Kometani, Mitsuhiro and Furukawa, Kenji and Yoneda, Takashi},
	date = {2021/12/13},
	doi = {10.1007/s11892-021-01423-2},
	id = {Nomura2021},
	isbn = {1539-0829},
	journal = {Current Diabetes Reports},
	number = {12},
	pages = {61},
	title = {Artificial Intelligence in Current Diabetes Management and Prediction},
	url = {https://doi.org/10.1007/s11892-021-01423-2},
	volume = {21},
	year = {2021}}

@article{mldm_2,
	author = {Lai, Hang and Huang, Huaxiong and Keshavjee, Karim and Guergachi, Aziz and Gao, Xin},
	date = {2019/10/15},
	doi = {10.1186/s12902-019-0436-6},
	id = {Lai2019},
	isbn = {1472-6823},
	journal = {BMC Endocrine Disorders},
	number = {1},
	pages = {101},
	title = {Predictive models for diabetes mellitus using machine learning techniques},
	url = {https://doi.org/10.1186/s12902-019-0436-6},
	volume = {19},
	year = {2019}}

@article{mldm_3,
	author = {Choi,Byoung Geol and Rha,Seung-Woon and Kim,Suhng Wook and Kang,Jun Hyuk and Park,Ji Young and Noh,Yung-Kyun},
	date = {2019/2/},
	isbn = {0513-5796},
	journal = {Yonsei Med J},
	month = {2},
	number = {2},
	pages = {191--199},
	publisher = {Yonsei University College of Medicine},
	title = {Machine Learning for the Prediction of New-Onset Diabetes Mellitus during 5-Year Follow-up in Non-Diabetic Patients with Cardiovascular Risks},
	url = {https://doi.org/10.3349/ymj.2019.60.2.191},
	volume = {60},
	year = {2019}}

@article{mldm_4,
	author = {Zhang, Liying and Wang, Yikang and Niu, Miaomiao and Wang, Chongjian and Wang, Zhenfei},
	date = {2020/03/10},
	doi = {10.1038/s41598-020-61123-x},
	id = {Zhang2020},
	isbn = {2045-2322},
	journal = {Scientific Reports},
	number = {1},
	pages = {4406},
	title = {Machine learning for characterizing risk of type 2 diabetes mellitus in a rural Chinese population: the Henan Rural Cohort Study},
	url = {https://doi.org/10.1038/s41598-020-61123-x},
	volume = {10},
	year = {2020}}

@article{vdb2021, title={On the Tractability of SHAP Explanations}, volume={35}, url={https://ojs.aaai.org/index.php/AAAI/article/view/16806}, DOI={10.1609/aaai.v35i7.16806}, abstractNote={SHAP explanations are a popular feature-attribution mechanism for explainable AI. They use game-theoretic notions to measure the influence of individual features on the prediction of a machine learning model. Despite a lot of recent interest from both academia and industry, it is not known whether SHAP explanations of common machine learning models can be computed efficiently. In this paper, we establish the complexity of computing the SHAP explanation in three important settings. First, we consider fully-factorized data distributions, and show that the complexity of computing the SHAP explanation is the same as the complexity of computing the expected value of the model. This fully-factorized setting is often used to simplify the SHAP computation, yet our results show that the computation can be intractable for commonly used models such as logistic regression. Going beyond fully-factorized distributions, we show that computing SHAP explanations is already intractable for a very simple setting: computing SHAP explanations of trivial classifiers over naive Bayes distributions. Finally, we show that even computing SHAP over the empirical distribution is #P-hard.}, number={7}, journal={Proceedings of the AAAI Conference on Artificial Intelligence}, author={Van den Broeck, Guy and Lykov, Anton and Schleich, Maximilian and Suciu, Dan}, year={2021}, month={May}, pages={6505-6513} }

@article{Horvitz01121952,
author = {D. G. Horvitz and D. J. Thompson},
title = {A Generalization of Sampling Without Replacement from a Finite Universe},
journal = {Journal of the American Statistical Association},
volume = {47},
number = {260},
pages = {663--685},
year = {1952},
publisher = {ASA Website},
doi = {10.1080/01621459.1952.10483446},


URL = { 
    
    
        https://www.tandfonline.com/doi/abs/10.1080/01621459.1952.10483446
    

},
eprint = { 
    
    
        https://www.tandfonline.com/doi/pdf/10.1080/01621459.1952.10483446
    

}

}

@article{biosppy,
    title = {BioSPPy: A Python toolbox for physiological signal processing},
    author = {Patrícia Bota and Rafael Silva and Carlos Carreiras and Ana Fred and Hugo Plácido {da Silva}},
    journal = {SoftwareX},
    volume = {26},
    pages = {101712},
    year = {2024},
    issn = {2352-7110},
    doi = {https://doi.org/10.1016/j.softx.2024.101712},
    url = {https://www.sciencedirect.com/science/article/pii/S2352711024000839},
}

@misc{puli2024explanationsrevealdefinitionencoding,
      title={Explanations that reveal all through the definition of encoding}, 
      author={Aahlad Puli and Nhi Nguyen and Rajesh Ranganath},
      year={2024},
      eprint={2411.02664},
      archivePrefix={arXiv},
      primaryClass={cs.LG},
      url={https://arxiv.org/abs/2411.02664}, 
}
